\documentclass[lettersize,journal]{IEEEtran}
\usepackage{amsmath,amsfonts}
\usepackage{algorithmic}
\usepackage{algorithm}
\usepackage{array}
\usepackage[caption=false,font=normalsize,labelfont=sf,textfont=sf]{subfig}
\usepackage{textcomp}
\usepackage{stfloats}
\usepackage{url}
\usepackage{verbatim}
\usepackage{graphicx}
\usepackage{cite}
\usepackage{multirow}
\usepackage{bbding}
\usepackage{graphicx}
\usepackage{booktabs}
\usepackage{xcolor}
\usepackage{colortbl}
\usepackage{hyperref}

\title{Boosting Segment Anything Model to Generalize Visually Non-Salient Scenarios}

\author{Guangqian Guo, Pengfei Chen, Yong Guo, Huafeng Chen, Boqiang Zhang, Shan Gao~\IEEEmembership{Member,~IEEE} \\
\thanks{Guangqian Guo, Huafeng Chen, and Shan Gao are with the Unmanned System Research Institute at Northwestern Polytechnical University, Xi'an 710072, China (e-mail: guogq21@mail.nwpu.edu.cn; chf@mail.nwpu.edu.cn; gaoshan@nwpu.edu.cn). Pengfei Chen is with the School of Electronic, Electrical, and Communication Engineering, University of Chinese Academic of Sciences, Beijing 101408, China (e-mail: chenpengfei20@mails.ucas.ac.cn). Yong Guo is with the Max Planck Institute for Informatics (MPI-INF) (e-mail: guoyongcs@gmail.com). Boqiang Zhang is with the University of Science and Technology of China (e-mail: cyril@mail.ustc.edu.cn).}
\thanks{This work was supported in part by the National Natural Science Foundation of China (NSFC) under Grant 62372382.}
}

\begin{document}

\maketitle
\footnotetext{
© 2025 IEEE. This is the author’s accepted manuscript.
Personal use of this material is permitted. Permission from IEEE must be obtained
for all other uses.
}

\begin{figure*}[h]
\centering
\includegraphics[width=0.99 \textwidth]{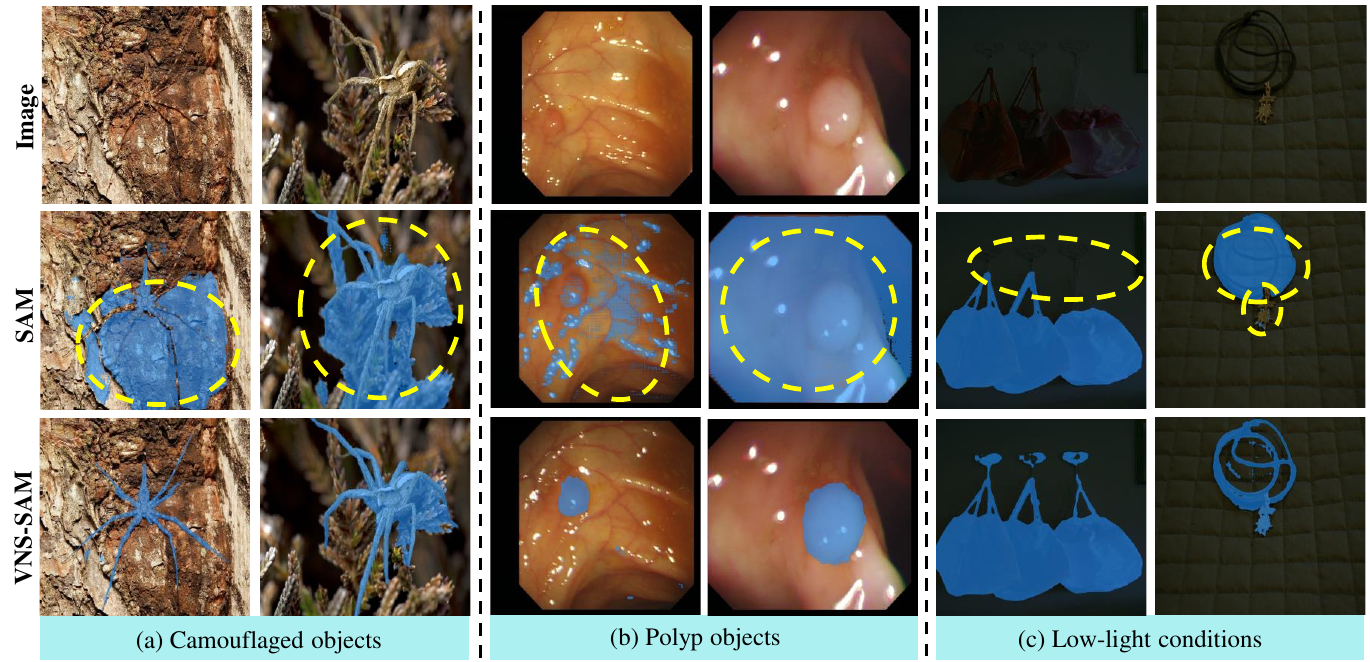}
\caption{A comparison of masks predicted by SAM and VNS-SAM under three typical non-salient scenarios. 
SAM (the second row) often struggles when dealing with (a) \textbf{camouflaged objects} where the objects perfectly match its surroundings, (b) \textbf{polyp objects} where polyp tissues and normal tissues have the same texture, posing challenges to medical image analysis, and (c) \textbf{objects in low-light conditions} where the targets lack significant color contrast with their backgrounds. SAM fails to accurately identify object boundaries and complete structures, leading to missing segmentation details and incorrect background predictions. In contrast, VNS-SAM (the third row) can produce more accurate segmentation. (Best viewed in color)
}
\vspace{-10pt}
\label{fig:failure-case}
\end{figure*}

\begin{abstract}
Segment Anything Model (SAM), known for its remarkable zero-shot segmentation capabilities, has garnered significant attention in the community. Nevertheless, its performance is challenged when dealing with what we refer to as visually non-salient scenarios, where there is low contrast between the foreground and background. 
{In these cases, existing methods often cannot capture accurate contours and fail to produce promising segmentation results.}
In this paper, we propose Visually Non-Salient SAM (VNS-SAM), aiming to enhance SAM's perception of visually non-salient scenarios while preserving its original zero-shot generalizability.    
We achieve this by effectively exploiting SAM's low-level features through two designs: Mask-Edge Token Interactive decoder and Non-Salient Feature Mining module.
These designs help the SAM decoder gain a deeper understanding of non-salient characteristics with only marginal parameter increments and computational requirements. The additional parameters of VNS-SAM can be optimized within 4 hours, demonstrating its feasibility and practicality.
In terms of data, we established VNS-SEG, a unified dataset for various VNS scenarios, with more than 35K images, in contrast to previous single-task adaptations. It is designed to make the model learn more robust VNS features and comprehensively benchmark the model's segmentation performance and generalizability on VNS scenarios. Extensive experiments across various VNS segmentation tasks demonstrate the superior performance of VNS-SAM, particularly under zero-shot settings, highlighting its potential for broad real-world applications.
Codes and datasets are publicly available at \href{https://guangqian-guo.github.io/VNS-SAM/}{{{https://guangqian-guo.github.io/VNS-SAM/}}}. 
\end{abstract}

\begin{IEEEkeywords}
Visually Non-Salient Characters, Segment Anything Model, Fine-tuning for Foundation Model
\end{IEEEkeywords}

\section{Introduction}
\label{sec:intro}
Accurate object segmentation~\cite{panoptic, yolact, videoseg, salient-detection, hanet, effective-rotate, hybridnet, alignment} in diverse scenarios is a fundamental task for various high-level visual applications. Recently, the Segment Anything Models (SAMs)~\cite{sam, sam2, sam3}, serving as a foundational segmentation model, has gained significant influence within the community due to its outstanding zero-shot segmentation capabilities. It can interactively segment any object in an image using visual prompts such as points and bounding boxes. Trained on the extensive SA-1B dataset, SAM's robust generalizability has led to breakthroughs and new paradigms in various downstream tasks, including remote sensing~\cite{rsprompter, sam-rs2, sam-rs3}, automatic data annotation~\cite{samrs, sam-label, p2p}, and medical image segmentation~\cite{sam-med, sam-med2}.

Some recent studies \cite{sam-struggle, sam-struggle2, sam-struggle-polyp} have pointed out that the performance of SAM decreases when faced with complex scenarios, such as camouflage, and polyps in medical images. 
{As shown in Fig. \ref{fig:failure-case}, due to the high intrinsic similarity between the foreground and background in VNS scenarios, SAM fails to effectively perceive subtle discriminative regions, confusing the foreground with the background, thus generating incorrect masks. 
This limitation severely restricts the applicability of SAM in the real world. Several studies \cite{polyp-sam, sam-med2d, medsam, samadapter, medical-sam-adapter} specialize the SAM for specific downstream tasks through fine-tuning and adapter modules. However, these methods only focus on a specific task, thereby overlooking the commonality knowledge across different complex scenarios and may compromise SAM's inherent generalization to other scenes. 
}

In this paper, we attempt to address this issue from a unified perspective. We found that some scenarios where SAM performs poorly share a common characteristic: low contrast between the foreground and background and blurred object boundaries (shown in Fig. \ref{fig:failure-case}). 
We refer to this commonality as \textbf{\textit{Visually Non-Saliency}} (VNS) and these scenarios as VNS scenarios. 
The unified perspective aims to jointly improve SAM's learning of the unified VNS characters thus consistently enhancing its performance in these VNS scenarios.  To learn the unified VNS knowledge, inspired by the fact that low-level features (such as edges and textures) are crucial for VNS object perception \cite{bfp, egnet, boundarycamo, edgepolyp, edge-detection-tip, edge-diffusion-tip, add_2, add_3}, we seek to effectively exploit them in SAM to boost its perception of VNS objects, which remains an open problem.

\begin{figure*}[t]
\centering
\includegraphics[width=0.95 \textwidth]{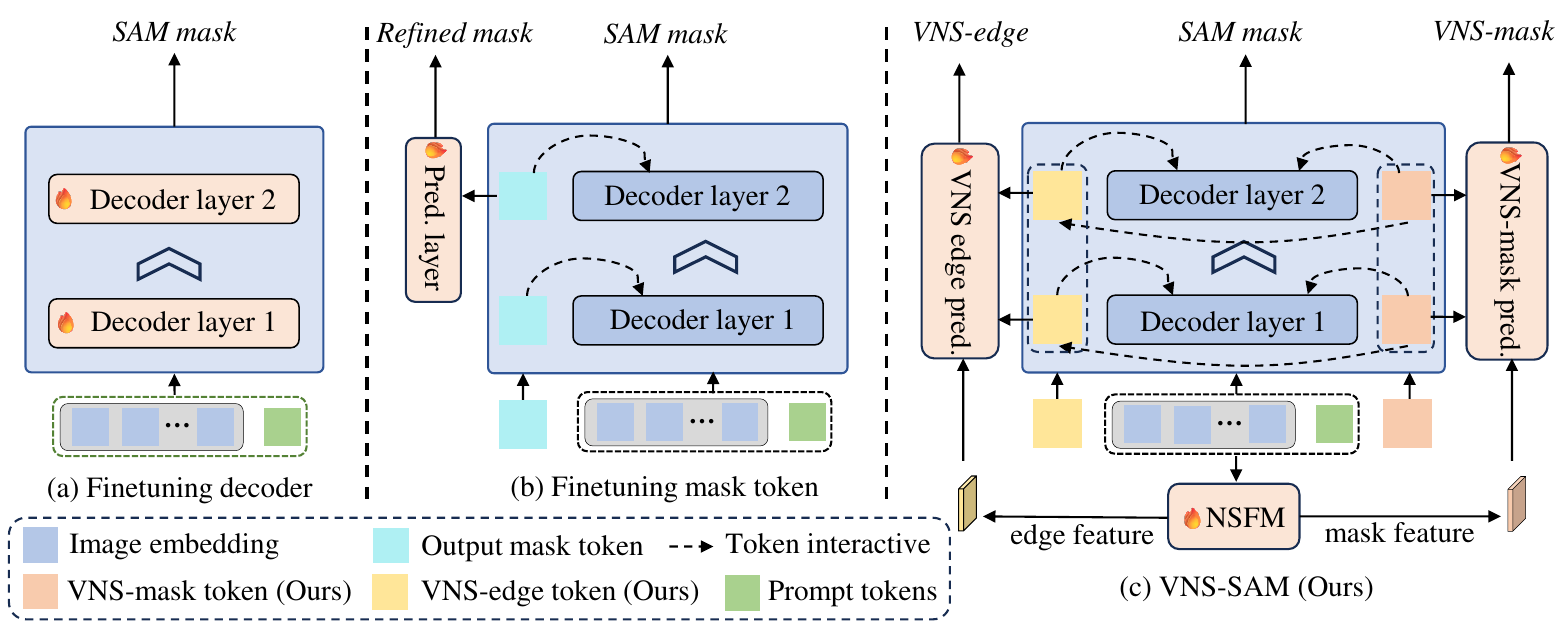}
\caption{ 
(a) Finetuning the entire decoder of SAM, (b) Finetuning additional output mask token to predict refined mask, (c) Our VNS-SAM integrates the interaction of edge semantics and dual-level decoder layers enhancement. The informative mask and edge features in the encoder are extracted by the Non-salient Feature Mining (NSFM) module, enriching the representation of the prediction layers for accurate segmentation. (Best viewed in color)
}
\vspace{-10pt}
\label{fig:compare}
\end{figure*}

To this end, we introduce \textbf{\textit{VNS-SAM}}, which effectively takes full advantage of SAM’s low-level features thus enhancing its perception of VNS characteristics by two key components.
First, we encourage SAM's decoder to efficiently learn VNS features by enhancing the perception of object edges. 
Instead of finetuning the entire mask decoder (Fig. \ref{fig:compare} (a)) or single mask token \cite{samhq} (Fig. \ref{fig:compare} (b)), we develop a mask-edge token interactive decoder (Fig. \ref{fig:compare} (c)). The core of this design is to enhance the mask prediction of SAM by introducing the interaction of VNS tokens (VNS-mask token and VNS-edge token) as well as dual-level enhancement to effectively boost the decoder to learn VNS characters. 
Second, we seek to mine the VNS features from the highly optimized image encoder to enrich the representation of the prediction layer. 
Accurate prediction of VNS objects requires fully exploring subtle discriminative features. To achieve this, we design a lightweight Non-Salient Feature Mining (NSFM) module to extract the most informative components from the SAM's encoder, thereby facilitating more precise predictions.
Built upon SAM, VNS-SAM leverages the generalization of the foundation model by freezing its original pre-trained parameters during the training stage. 
Note that the proposed VNS-SAM only brings {9.8 M} parameters and can be trained efficiently within 4 hours on 4$\times$ RTX 4090 GPUs.

In terms of the data, to enable the model to learn the VNS characters, instead of adapting to a single dataset, we establish a unified dataset for VNS scenarios, named \textbf{\textit{VNS-SEG}}.  
This not only benefits the model in learning more robust non-salient features but also improves the model's performance across multiple tasks. 
VNS-SEG comprises 35K image-mask pairs with diverse VNS scenarios, sourced from the well-known existing datasets and our synthesized data.
The training set of VNS-SEG consists of 23,232 images and the evaluation set comprises 11 subsets across 4 VNS scenarios. 
The evaluation set is divided into the seen-set and unseen-set, for comprehensively benchmarking the zero-shot transfer ability of models. 
We hope the constructed VNS-SEG dataset will inspire more segmentation models suitable for VNS scenarios and be valuable for future research.

Overall, the major contributions of this work can be summarized in four aspects. 

\begin{itemize}
    \item We analyze SAM's limitations in a series of scenarios with low contrast between the foreground and background, which we collectively refer to as VNS scenarios.
    Thus, we propose VNS-SAM, a generalized interactive segmentation model built upon SAM, with improved robustness against various VNS scenarios. 
    \item We develop a Mask-Edge Token Interactive decoder and a Non-Salient Feature Mining module in VNS-SAM to encourage the model to mine subtle discriminative features. The proposed method brings negligible parameters and can be trained efficiently in less than 4 hours.
    \item  We constructed a unified dataset, VNS-SEG, comprising more than 35K image-mask pairs for training and evaluating the model optimally. Compared to single-task datasets, this unified dataset benefits the model in learning more robust VNS characters. VNS-SEG aims to establish a new benchmark for VNS segmentation.
    \item We conduct extensive experiments and the results show that VNS-SAM achieves superior segmentation performance on various VNS scenarios and retains powerful interactive segmentation generalizability. 
\end{itemize}

\section{Related Work}    

\textbf{{Segment Anything Model and Variants.}} 
Segment Anything Model (SAM)~\cite{sam, sam2} has gained significant influence within the community due to its outstanding zero-shot segmentation capabilities. 
Serving as a foundational segmentation model, SAM is trained on an extensive SA-1B dataset \cite{sam}, consisting of over 11 million images and one billion masks. SAM can interactively segment any object in an image using prompts such as points and bounding boxes.
Its robust generalization abilities have led to breakthroughs and new paradigms in various downstream tasks \cite{add_1, sam-label, samrs, sam-wsss, sam-wsss2, samantic-aware-sam, sam-cod, p2p, glesam}.

Although SAM is powerful, its performance decreases when facing complex real-world scenarios, such as objects with intricate structures \cite{samhq} or camouflaged objects \cite{sam-struggle, sam-struggle2, sam-not-perfect}. 
Enhancing SAM's capability in such challenging scenarios is a worthwhile research topic. 
Based on SAM, some improved variants have been proposed, which can be roughly categorized into two routes. 
One route  \cite{medical-sam-adapter, samadapter, rsprompter, polyp-sam} involves using SAM for specific downstream tasks through domain-specific finetuning.
These efforts typically focus on improving SAM's performance on a specific task or dataset while sacrificing the model’s inherent generalization capabilities.           
Another route \cite{samhq, robustsam, asam, mobilesam, efficientsam, glesam} is to extend SAM's capabilities, preserving its strong generalization performance. 
For example, MobileSAM~\cite{mobilesam} and EfficientSAM~\cite{efficientsam}, through techniques like knowledge distillation, make it applicable to real-time segmentation. ASAM \cite{asam} enhances SAM's generalization capabilities through adversarial training.
HQ-SAM~\cite{samhq} has improved SAM's segmentation quality for objects with complex structures by adding adaption layers while freezing SAM's original parameters.
Diverging from these existing methods, our method aims to enhance the segmentation capability of SAM in visually non-salient scenarios from a unified perspective while preserving its generalization abilities. 

\textbf{{Object Segmentation in Visually Non-Salient Scenarios.}}
Unlike general scenarios, there are some challenging scenes in the real world where the foreground and background of objects have similar textures and colors, making the objects difficult to detect.
We refer to the scenarios with this character as visually non-salient (VNS) scenarios, \textit{e.g.}, camouflaged scenarios \cite{cod10k, camo, nc4k, csu, fspnet, sinetv2, fder, camo-frequency, focusdiffuser, pcod} and polyp tissues in medical images \cite{etis, CVC-ClinicDB, ColonDB, kvasir, edgepolyp, pranet}, and low-light environments \cite{lis, featenhancer, lol}.
Accurate perception and understanding of these VNS scenarios remain a challenging issue. Some related works usually design task-specific model structures, such as feature encoders and mask decoders to solve one specific task.    
For example, in the camouflaged object detection task, SINet \cite{cod10k} designs a bio-inspired network to gradually search and locate the camouflaged object. 
In the medical domain for polyp segmentation, PraNet \cite{pranet} integrates the Reverse Attention module to accentuate the boundaries between polyps and their surroundings.
However, these methods and the datasets they use are task-specific (one model solves one task). 
Different from the existing works, we seek to solve this problem from another perspective. 
We constructed a unified non-salient dataset to enable the model to effectively learn more robust VNS characters. 
Furthermore, building upon SAM, we develop a general segmentation model that achieves superior performance in several non-salient scenarios while preserving powerful generalization capability.

\textbf{{Edge-boosted Segmentation Methods.}}
Many segmentation methods introduce effective low-level features (such as edge information) into the network to enhance the model's perception of local details, thereby improving the segmentation capability of the model~\cite{bfp, basnet, egnet, boundarycamo,  edgepolyp, boundary-medical, tip-edge, tip-edge2}.  
The core of these methods lies in designing an edge-aware module to capture richer context and detailed information, contributing to accurate segmentation. For instance,
Zhou \textit{et al.}~\cite{boundarycamo} designed a boundary guidance module to learn boundary-enhanced feature representations for camouflaged object detection. In this paper, we propose to exploit low-level features in SAM and encourage SAM’s decoder to efficiently learn VNS features by enhancing the perception of object edges, thereby boosting the segmentation in VNS scenarios.

\begin{figure*}[t]
\centering
\includegraphics[width=1. \textwidth]{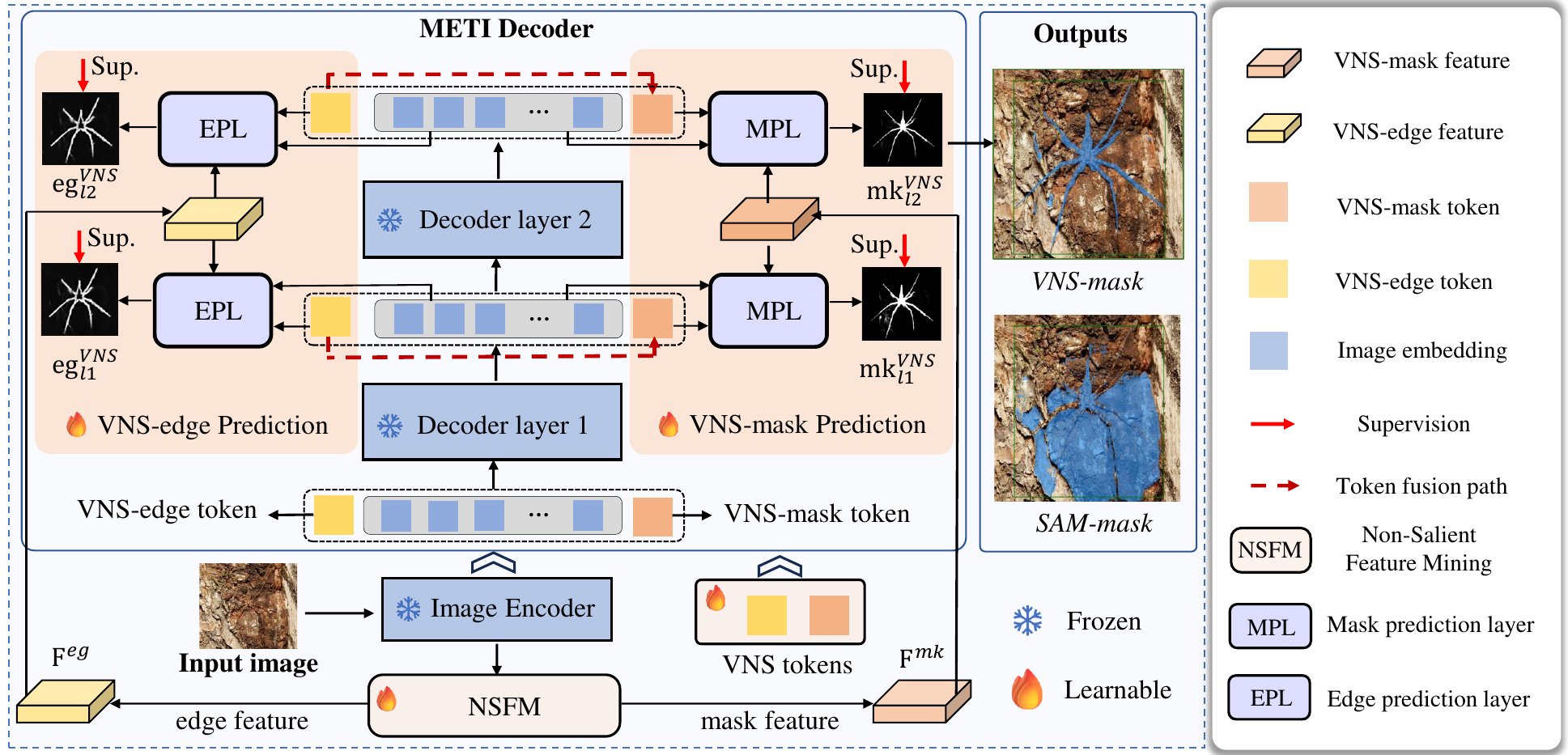}
\caption{
{Overview. Building upon SAM, VNS-SAM makes two key improvements. 
First, it enhances the SAM's original decoder to a mask-edge token interactive (METI) decoder with the interaction of edge semantics and dual-level decoder layers enhancement. 
Second, a lightweight non-salient feature mining (NSFM) module is designed to extract mask and edge features from the image encoder to enrich the representation of the mask and edge prediction layers (\textit{i.e.}, MPL and EPL). 
During training, the parameters of the pre-trained SAM are frozen, with the newly added parameters in VNS-SAM trained. 
During inference, VNS-SAM outputs the more precise VNS-mask and the original SAM-mask.
The prompt encoder and prompt tokens are omitted here.
}}
\label{fig:framework}
\end{figure*}

\section{{VNS-SAM:} Visually Non-Salient Segment Anything Model}

In the following, we focus on improving the segmentation quality of SAM in visually non-salient (VNS) scenarios and develop a more powerful generalized segmentation model.
{To get started, we analyze SAM’s limitations in a series of scenarios with low contrast between the foreground and background, which we collectively refer to as {\textbf{\textit{Visually Non-salient Scenarios}}} in Section \ref{analysis}.
We highlight that these scenarios are quite common and the poor performance greatly limits SAM's realistic applications.
To address these issues, we propose two novel techniques to boost the SAM to generalize VNS scenarios.}
First, in Section \ref{sec: meti}, we propose a \textbf{\textit{Mask-Edge Token Interactive Decoder}} that encourages the SAM's decoder to explicitly perceive useful non-salient information. This is achieved by using a pair of learnable VNS-tokens and dual-level enhancement as illustrated in Fig. \ref{fig:framework}. Second, in Section \ref{tfd}, we develop a \textbf{\textit{Non-Salient Feature Mining}} module to enrich the feature representation for improving the quality of VNS-mask predictions.
Both methods are lightweight, and we will show that they greatly improve the segmentation performance of SAM in VNS scenarios.

\vspace{-2mm}
\subsection{{Visually Non-Salient Scenarios and Limitations of SAM}}
\label{analysis}
Unlike general scenes, there are many challenging scenarios in the real world, where the foregrounds and backgrounds have low contrast and similar textures and colors, making the target objects difficult to perceive precisely. 
For example, as illustrated in Fig. \ref{fig:failure-case}, camouflaged objects are extremely similar to their surroundings, making them hard to prey on by their natural enemies. In medical images, polyps and normal tissues have the same texture and are mostly small in shape, posing challenges to medical image analysis. 
Additionally, objects in low-light conditions lack significant color contrast with their backgrounds. 
In this paper, \textbf{\textit{we collectively refer to the characters of such scenes as visually non-salient (VNS) characters, and the scenarios with VNS characters are termed VNS scenarios.}} 
Due to a lack of ability to extract VNS features and the absence of the corresponding dataset for training, SAM generally performs poorly in VNS scenarios. 

As illustrated in Fig. \ref{fig:failure-case}, we can find that SAM struggles to perceive the foreground of the VNS objects, resulting in incorrect segmentation. This indicates the weak robustness of SAM in VNS scenarios.
To address this, different from the previous methods, we seek to consistently enhance SAM's segmentation ability in various VNS scenarios while retaining its original generalizability. We achieve this by designing two effective techniques in the remainder of this section.

\subsection{Mask-Edge Token Interactive Decoder}
\label{sec: meti}
In the first part of our method, we seek to encourage SAM's decoder to learn more about VNS characteristics. 
{Some previous methods ~\cite{bfp, egnet, boundarycamo, edgepolyp, mambafusion} proved that extracting and learning low-level features (such as edges and local details) are crucial for VNS object perception.} This motivates us to fully exploit SAM's low-level features to enhance the perception of less discriminative characteristics, which has rarely been studied.  
To achieve this, we incorporate edge semantics into the SAM decoder. Specifically, we develop a {Mask-Edge Token Interactive (METI)} decoder by introducing a pair of Visually Non-Salient Tokens and Dual-level Prediction Enhancement. 
The detailed structure is illustrated in Fig. \ref{fig:framework}.

\textbf{Visually Non-Salient Tokens.}
{We add a pair of VNS-tokens that contain a VNS-mask token  $\mathbf{e}^{mk} \in \mathbb{R}^{1 \times 256}$ and a VNS-edge token $\mathbf{e}^{eg} \in \mathbb{R}^{1 \times 256}$ into SAM's decoder.} 
By reusing the SAM's original layer, the VNS-tokens are concatenated with the original pre-trained output tokens and prompt tokens. Then, these tokens together with image embeddings, defined as $\{\mathbf{e}^{sam},\mathbf{e}^{mk}, \mathbf{e}^{eg}\}$, are fed into the mask decoder. 
In each decoder layer, we reuse the two-way transformer block in the original mask decoder to interact features among tokens and between tokens and image embeddings, respectively. 
\begin{equation}
{\mathbf{F}}, \{{\mathbf{e}}^{sam},{\mathbf{e}}^{mk}, {\mathbf{e}}^{eg} \} \gets \Phi_\text{twt}(\Phi_\text{twt}(\mathbf{F}, \{\mathbf{e}^{sam},\mathbf{e}^{mk}, \mathbf{e}^{eg}\})),
\label{eq: twt2}
\end{equation} 
where ${\mathbf{F}}$ and $\{{\mathbf{e}}^{sam},{\mathbf{e}}^{mk}, {\mathbf{e}}^{eg}\}$ on the left side denote the embedding features and tokens updated after two decoder layers, respectively. $\Phi_\text{twt}(\cdot)$ indicates the two-way transformer layer (containing a self-attention unit and an image-to-token and token-to-image attention block). 
{During training, the VNS-edge token serves as a low-level feature learner, providing effective low-level information to the VNS-mask token. Specifically, in the decoding process, the interaction between VNS-tokens occurs in two ways. On the one hand, the VNS-edge token implicitly interacts with the mask token via the cross-attention mechanism in the original decoder layer, propagating effective edge and texture information to the mask token. Additionally, we explicitly strengthen the interaction between the two tokens through a straightforward fusion operation. After each decoder layer, the VNS-edge token is also explicitly integrated with the VNS-mask token, as}
\begin{equation}
{\mathbf{e}}^{mk} \gets \mathcal{F}_{token}({\mathbf{e}}^{mk},  {\mathbf{e}}^{eg}).
\label{eq: tokenfusion}
\end{equation} 
$\mathcal{F}_{token}(\cdot)$ denotes the token integration operations. The two tokens first perform element-wise addition, followed by fusion through a linear layer.
{This simple operation further enables the explicit aggregation of edge representation from the VNS-edge token to the VNS-mask token.}

\begin{figure}[t]
\centering
\includegraphics[width=0.49 \textwidth]{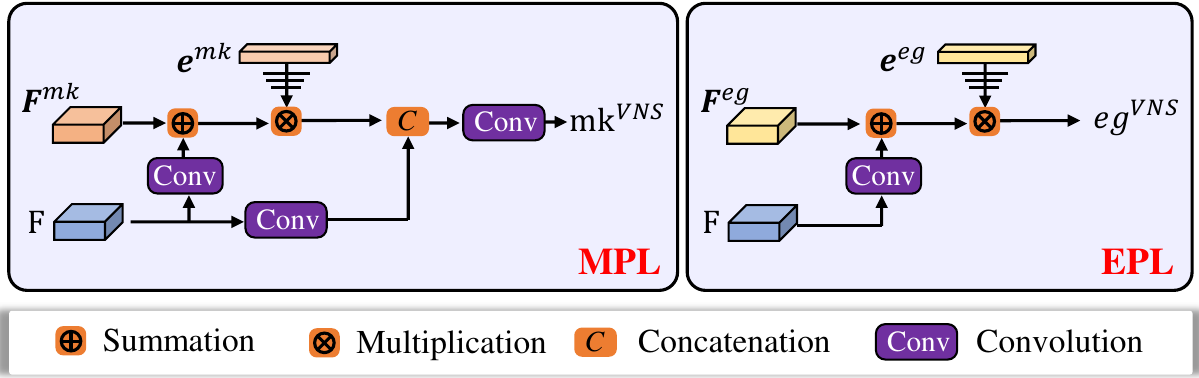}
\caption{
{Details of Mask Prediction Layer (MPL) and Edge Prediction Layer (EPL). In MPL, we reuse the highly optimized image embeddings $\text{F}$ as supplementary features. (Best viewed in color)
}
}
\label{fig:mpl-epl}
\end{figure}

\begin{figure}[t]
\centering
\includegraphics[width=0.49 \textwidth]{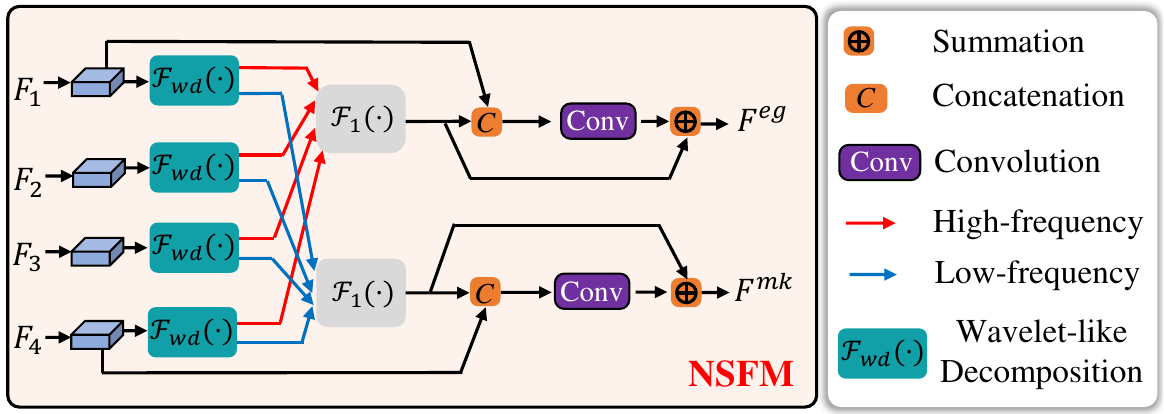}
\caption{Details of Non-Salient Feature Mining (NSFM) module. The multi-level features extracted from the backbone are decomposed into different components. Then, the most informative high-frequency and low-frequency components are selected and multi-level features are aggregated for edge and mask feature extraction. (Best viewed in color)}
\label{fig:nsfm}
\end{figure}

\textbf{Dual-level Prediction Enhancement.}
{To allow both decoding layers to consistently learn subtle discriminative features, thereby improving the decoder's ability to understand non-salient characters, we improve the original decoder to a dual-level enhanced decoder by hierarchical supervision and prediction enhancement. 
\textbf{Hierarchical supervision:} After each decoder layer, the interacted VNS-mask and VNS-edge tokens are input to the enhanced Mask Prediction Layer (MPL) and Edge Prediction Layer (EPL) to obtain the mask and edge predictions. During the training stage, mask and edge predictions from both decoder layers are supervised by ground-truths.
\textbf{Prediction enhancement:} As shown in Fig. \ref{fig:mpl-epl}, in both MPL and EPL, the VNS-mask feature ${F}^{mk}$ and VNS-edge feature ${F}^{eg}$ are first fused respectively with the image embedding $\mathbf{F}$. Then, each VNS token ($\mathbf{e}^{mk}$ and $\mathbf{e}^{eg}$) passes through a learnable MLP layer and then performs a dot product with the corresponding fused feature to obtain a single-channel output that is used as the edge prediction in EPL.
Additionally, in MPL, to further utilize the highly optimized image embedding $\mathbf{F}$ in SAM, we pass it to two learnable convolutional layers to obtain a single-channel feature map as the supplementary feature that has the same size as the output by the VNS-mask token. Finally, these two feature maps are concatenated and passed through two convolutional layers with a kernel of 3$\times$3 and 1$\times$1 to fuse them and output the final mask prediction.}

\subsection{Non-salient Feature Mining}
\label{tfd}
In the second part of our method, we seek to mine the useful low-level features from the highly-optimized image encoder to enrich the representation of the prediction layer.  
Based on this objective, we propose a learnable NSFM module shown in Fig. \ref{fig:nsfm}, which effectively extracts the VNS mask and edge representations from the multi-level image encoder. Guided by the biological study, 
discriminative features mainly exist in the high-frequency and low-frequency components of features \cite{animalcamou}. Thus, we first decompose the extracted features to obtain different components. Then we select and aggregate the most informative components for further mask and edge extraction. Note that the proposed module is lightweight, bringing only about {3M} parameter increase.

Specifically, given multi-level features extracted by the image encoder, {we adopt the Haar discrete wavelet decomposition \cite{haar} that is mathematically rigorous and widely used in feature analysis and segmentation \cite{wavelet-sffnet, wavelet-polyp} for decomposing multi-level features.
To be specific, the Haar discrete wavelet decomposes each feature into four wavelet sub-bands, $F_{k}^{HH}, F_{k}^{HL}, F_{k}^{LH}, F_{k}^{LL}$.
\begin{equation}
F_{k}^{HH}, F_{k}^{HL}, F_{k}^{LH}, F_{k}^{LL} = \mathcal{F}_{wd}(F_{k}), k = 1,2,3,4,
\label{eq: wd}
\end{equation} 
where $\mathcal{F}_{wd}(\cdot)$ denotes the Haar wavelet decomposition.
We select the most informative high-frequency $F_{k}^{HH}$ and low-frequency $F_{k}^{LL}$ components. 
}
Then, multi-level high-frequency and low-frequency components are respectively aggregated to obtain enhanced representations $F_{agg}^{HH}$ and $F_{agg}^{LL}$. 
Taking the high-frequency components as an example, the high-frequency features of multi-levels are first concatenated and passed through a 1$\times$1 convolution layer for channel reduction. 
Then, effective attention layers \cite{ca, sa} are applied to explore the inter-layer feature correlations. 
After that, we obtain the multi-level integrated high-frequency features. 
The above operations can be denoted as: 
\begin{equation}
F_{agg}^{HH} = \mathcal{F}_1(\{F_k\}_{k=1}^{4}) = \text{Attn}(\text{Conv}(\text{Cat}(\{F_k^{HH}\}_{k=1}^{4}))),
\label{eq: agg}
\end{equation} 
where $\text{Attn}(\cdot)$ indicates the attention layer that joins channel and spatial attention layers. 
Similarly, the aggregated low-frequency feature $F_{agg}^{LL}$ can be obtained.

As previously analyzed, the high-frequency components contain rich texture and edge information, thus we use them to extract visually non-salient edge features, while the low-frequency components extract visually non-salient mask features. For the high-frequency part, the shallow layer $F_1$ is concatenated with $F_{agg}^{HH}$ as supplementary information, and then a 1$\times$1 convolution layer is used to fuse the concatenated features and reduce the channel dimension. {Finally, a skip connection is used to merge the high-frequency components and the supplemented representation to generate the VNS-edge feature $F^{eg}$. Similarly, we can obtain the VNS-mask feature $F^{mk}$.}
The VNS-mask and VNS-edge features are used to make mask and edge predictions in MPL and EPL, respectively, as stated in the above part.

\begin{table*}[t!]
\centering
\renewcommand\arraystretch{1.2}
\setlength{\abovecaptionskip}{0pt}%
\setlength{\belowcaptionskip}{5pt}%
\caption{Data composition of the training set of our VNS-SEG. It comprises a total of 23,232 images that are sourced from renowned existing datasets and synthesized data. The collected images contain diverse visually non-salient characters.}
\setlength{\tabcolsep}{2.9 mm}{
\begin{tabular}{c|c|c|c|c|c|c|c|c}\toprule[1.pt]
  \multirow{2}{*}{\textit{Train set}} & \multicolumn{4}{c|}{ Existing Datasets} & \multicolumn{3}{c|}{ Synthesized Datasets} & \multirow{2}{*}{Sum} \\ \cline{2-8}
  &  CAMO \cite{camo} & COD10K \cite{cod10k} & Kvasir \cite{kvasir} & Clin.DB \cite{CVC-ClinicDB}  & DIS-Dark \cite{dis} & Thin-Dark \cite{thin} & FSS-Dark \cite{fss} &  \\
\toprule[1.pt]
 Number & 1000 & 3040 &900 & 550& 3000 & 4742 & 10000 & 23232\\
\toprule[1.pt]
\end{tabular}}
\label{tab:tra}
\end{table*}

\begin{table*}[htp]
\centering
\renewcommand\arraystretch{1.2}
\setlength{\abovecaptionskip}{0pt}%
\setlength{\belowcaptionskip}{5pt}%
\caption{Data composition of the evaluation set for VNS-SEG, comprising 11 subsets across 4 VNS scenarios. It is divided into seen-set and unseen-set to fully evaluate the model's segmentation performance and generalization ability in VNS scenarios.  Note that all data in the unseen set are collected from real-world scenarios, ensuring an effective evaluation of VNS-SAM's performance in realistic applications.
}
\setlength{\tabcolsep}{3.7 mm}{
\begin{tabular}{c|c|c|c|c|c|c|c}\toprule[1.pt]
 {\textit{Eval-Seen-Set}} &  CAMO \cite{camo} & COD10K \cite{cod10k} & Kvasir \cite{kvasir} & ClinicDB \cite{CVC-ClinicDB}  & DIS-Dark \cite{dis} & Thin-Dark \cite{thin} & \multirow{2}{*}{Sum} \\ \cline{1-7}
 Number & 250 & 2026 &100 & 62 & 480 & 1000 &  \\ 
 \toprule[0.9pt]
 {\textit{Eval-Unseen-Set }}& {NC4K \cite{nc4k}}   & ColonDB \cite{ColonDB} & ETIS \cite{etis}  &  LIS \cite{lis} & CDS2K \cite{csu}  & -  & \multirow{2}{*}{12175}\\ \cline{1-7}
 Number & {4121} & 380 & 196 & 2230 & 1330 & - \\
\toprule[1.pt]
\end{tabular}}
\label{tab:eval-set}
\vspace{-1mm}
\end{table*}

\subsection{Training and Inference}
\label{sec: train}

\textbf{{Training.}}
During training, we freeze the pre-trained SAM's weights and only update the parameters in the newly added modules. 
{We use a mixture of sampled prompts, including bounding boxes, randomly selected points, and coarse masks.}
The images and prompts are fed into VNS-SAM and generate two levels of mask and edge predictions in the decoder. 
For the mask supervision, we employ the structure loss $\mathcal{L}^{stru}$ \cite{structure-loss} that contains the weighted IoU loss and the weighted binary cross-entropy loss. It focuses more on hard pixels.  
For edge supervision, we use the dice loss $\mathcal{L}^{dice}$ \cite{diceloss}.
The total loss is formulated as:
\begin{equation}
\mathcal{L}_{total} = \sum_{k=1}^{2}  \mathcal{L}_{l_k}^{stru}(\mathbf{mk}^{VNS}_{l_k}, \mathbf{mk}^{gt}) + \sum_{k=1}^{2} \mathcal{L}^{dice}_{l_k}(\mathbf{eg}^{VNS}_{l_k},\mathbf{eg}^{gt}).
\end{equation}

\textbf{{Inference.}}
During the inference phase, we discard the output of the edge token and the first layer mask output. 
Only the VNS mask of the second decoder layer and the original SAM's output are computed. 
We up-sample the predicted masks to the original image's resolution as the final output.

\begin{figure}[t]
\centering
\includegraphics[width=0.49 \textwidth]{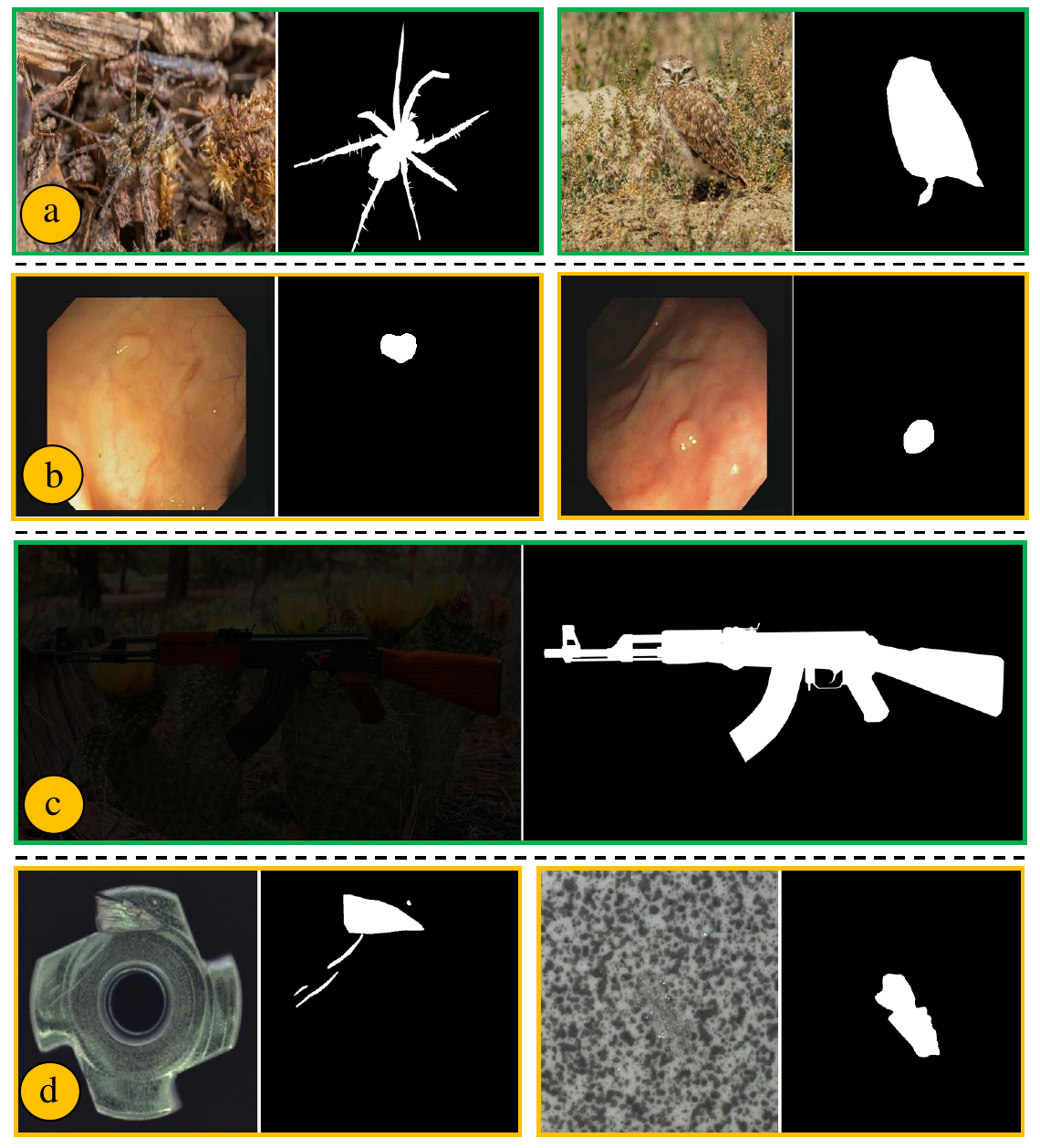}
\caption{Examples of images and corresponding masks in VNS-SEG that contain diverse VNS scenarios, \textit{i.e.}, (a) camouflaged objects, (b) polyp tissues, (c) objects in low-light conditions, and (d) industrial defects. 
It can be visualized that these objects have high similarity with backgrounds, making them difficult to perceive and challenging the current segmentation models.
}
\label{fig:data-vis}
\end{figure}

\section{{VNS-SEG: Visually Non-salient Segmentation Dataset}}
\label{dataset}
To enable the segmentation models to effectively learn VNS characters, we meticulously construct a unified dataset: VNS-SEG, for training and benchmarking the performance of the segmentation model on diverse VNS scenarios. 
It contains more than 35K image-mask pairs. The images in VNS-SEG are sourced from some well-known existing datasets in the community and synthesized data to enrich the diversity of our dataset.
The unified dataset allows the model to learn more robust non-salient characters and improves the performance across multiple VNS tasks. Fig. \ref{fig:data-vis} shows randomly selected images and corresponding mask annotations from VNS-SEG. 
Compared to normal objects, objects in VNS scenarios are harder to observe, greatly increasing the difficulty of precise segmentation and challenging the current model.

\subsection{Dataset Construction}
\textbf{Training Set.} 
The training set contains 23,232 images with accurate mask annotations. To construct the training set, we carefully select four well-known datasets in the community that have challenging VNS characters, including COD10K~\cite{cod10k} and CAMO~\cite{camo}, Kvasir~\cite{kvasir}, and ClinicDB~\cite{CVC-ClinicDB}.
Specifically, COD10K and CAMO are from camouflaged datasets, containing 3040 and 1000 images, respectively.  Kvasir and ClinicDB are from polyp segmentation datasets that comprise 900 and 550 images, respectively. 
Additionally, to enrich the diversity of the data, we synthesize some images under low-light conditions. 
We achieve this by training a CycleGAN~\cite{cyclegan} model that transforms normal images into low-light images. 
We select DIS ~\cite{dis}, ThinObject-5K~\cite{thin}, and FSS~\cite{fss} datasets, characterized by fine-grained details and complex geometries, as source datasets, and transform them into low-light datasets, \textit{i.e.}, DIS-Dark, Thin-Dark, and FSS-Dark. 
By doing so,  we can directly utilize the precise mask annotations of the original datasets, and the synthesized data can significantly enrich our dataset's diversity.
The composition details are shown in Tab. \ref{tab:tra}.

\begin{table*}[t]
\centering
\renewcommand\arraystretch{1.25}
\setlength{\abovecaptionskip}{0pt}%
\setlength{\belowcaptionskip}{5pt}%
\caption{{Performance comparison on the \textit{\textbf{Eval-Seen-Set}} of VNS-SEG. Three types of prompts are used. We finetuned HQ-SAM on our VNS-SEG dataset, termed HQ-SAM-F. Our models consistently outperform the baseline SAM and other competitors on diverse subdatasets and prompts. The words with boldface indicate the best results.}
}
\setlength{\tabcolsep}{0.58 mm}{
\begin{tabular}{c|cccc|cccc|cccc|cccc|cccc|cccc}\toprule[1.pt]
     \multirow{2}{*}{{Method}} & \multicolumn{4}{c|}{{CAMO}} & \multicolumn{4}{c|}{{COD 10K}} &\multicolumn{4}{c|}{{Kvasir}}  & \multicolumn{4}{c|}{ClinicDB} & \multicolumn{4}{c|}{DIS-Dark} & \multicolumn{4}{c}{Thin-Dark}\\
\cline{2-25}
                                & IoU & BIoU & $E_{\phi}$ & $F_{\beta}^{w} $    & IoU & BIoU   & $E_{\phi}$ & $F_{\beta}^{w} $  & IoU & BIoU     &  $E_{\phi}$ & $F_{\beta}^{w} $  & IoU & BIoU     &  $E_{\phi}$ & $F_{\beta}^{w} $  & IoU & BIoU     &  $E_{\phi}$ & $F_{\beta}^{w} $  & IoU & BIoU     &  $E_{\phi}$ & $F_{\beta}^{w} $ 
                                \\
\toprule[1.pt]
\multicolumn{25}{c}{\textit{\textbf{Point-Prompting-Based Evaluation}}} \\
\hline
SAM \cite{sam} 
& .659   & .450   & .757   & .633
& .687   & .593   & .830   & .665 
& .719   & .480   & .800   & .652
& .602   & .447   & .753   & .528
& .594   & .476   & .723   & .558
& .775   & .617   & .827   & .752
\\
HQ-SAM \cite{samhq}
& .769   & .522   & .886   & .790
& .771   & .672   & .922   & .803 
& .799   & .568   & .883   & .807 
& .727   & .560   & .851   & .735 
& .721   & .620   & .872   & .752 
& .853   & .714   & .916   & .867  \\
HQ-SAM-F \cite{samhq}
& .780    & .542   & .880   & .769 
& .774    & .675   & .911   & .771 
& .782   & .554   & .856   & .695 
& .690   & .500   & .828   & .634 
& .720   & .624   & .861   & .723 
& .854   & .718   & .905   & .840 \\

\rowcolor{lightgray!40} 
VNS-SAM (Ours)
& \textbf{.797}   & \textbf{.562}   & \textbf{.917}   & \textbf{.830} 
& \textbf{.813}   & \textbf{.723}   & \textbf{.953}   & \textbf{.854} 
& \textbf{.877}   & \textbf{.666}   & \textbf{.949}   & \textbf{.902} 
& \textbf{.833}   & \textbf{.662}   & \textbf{.956}  & \textbf{.878} 
& \textbf{.780}   & \textbf{.705}   & \textbf{.922}   & \textbf{.815} 
& \textbf{.889}   & \textbf{.777}   & \textbf{.947}   & \textbf{.906} \\
\hline
\multicolumn{25}{c}{\textit{\textbf{Noise-Box-Prompting-Based Evaluation}}} \\
\hline
SAM \cite{sam}
& .740   & .510   & .852   & .765
& .725   & .629   & .879   & .751
& .756   & .524   & .853   & .784
& .828   & .648   & .944   & .859
& .529   & .434   & .695   & .545
& .656   & .514   & .716   & .646 \\
HQ-SAM \cite{samhq} 
& .756   & .517   & .890   & .801 
& .747   & .646   & .913   & .785 
& .802   & .585   & .890   & .824 
& .821   & .655   & .947   & .859 
& .682   & .583   & .845   & .715 
& .799   & .653   & .870   & .801  \\
HQ-SAM-F \cite{samhq}
& .761   & .531   & .890   & .803 
& .773   & .679   & .925   & .804 
& .824   & .612   & .909   & .843 
& .836   & .678   & .957   & .868 
& .708   & .620   & .865   & .738 
& .823   & .682   & .888   & .824  \\
\rowcolor{lightgray!40} 
VNS-SAM (Ours)
& \textbf{.769}   & \textbf{.543}   & \textbf{.904}   & \textbf{.816}
& \textbf{.785}   & \textbf{.695}   & \textbf{.939}   & \textbf{.825} 
& \textbf{.826}   & \textbf{.623}   & \textbf{.918}   & \textbf{.852} 
& \textbf{.842}   & \textbf{.682}   & \textbf{.964}   & \textbf{.879} 
& \textbf{.740}   & \textbf{.670}   & \textbf{.902}   & \textbf{.786} 
& \textbf{.845}   & \textbf{.724}   & \textbf{.916}   & \textbf{.860}  \\
\hline
\multicolumn{25}{c}{\textit{\textbf{GT-Box-Prompting-Based Evaluation}}} \\
\hline
SAM \cite{sam} 
& .757   & .532   & .864   & .779
& .746   & .650   & .892   & .771
& .762   & .534   & .861   & .798
& .818   & .652   & .946   & .862
& .579   & .475   & .722   & .584
& .652   & .516   & .689   & .631
\\
HQ-SAM \cite{samhq} 
& .776   & .535   & .898   & .812 
& .757   & .658   & .918   & .794
& .810   & .594   & .888   & .829 
& .825   & .671   & .946   & .859 
& .703   & .605   & .854   & .729 
& .810   & .672   & .870   & .807  \\
HQ-SAM-F \cite{samhq} & .783   & .553   & .902   & .819 
& .785   & .692   & .930   & .813 
& .837   & .627   & .910   & .849 
& .842   & .686   & .956   & .873 
& .727   & .637   & .872   & .752 
& .838   & .705   & .894   & .834 \\
\rowcolor{lightgray!40} 
VNS-SAM (Ours)
& \textbf{.795}   & \textbf{.575}   & \textbf{.916}   & \textbf{.834}
& \textbf{.800}   & \textbf{.715}   & \textbf{.944}   & \textbf{.838} 
& \textbf{.848}   & \textbf{.659}   & \textbf{.919}   & \textbf{.866} 
& \textbf{.851}   & \textbf{.698}   & \textbf{.963}   & \textbf{.886} 
& \textbf{.765}   & \textbf{.698}   & \textbf{.911}   & \textbf{.802} 
& \textbf{.868}   & \textbf{.757}   & \textbf{.923}   & \textbf{.874}  \\
\toprule[1.pt]
\end{tabular}}
\label{tab:seen-set}
\vspace{-3mm}
\end{table*}

\textbf{Evaluation Set.}
The evaluation set consists of 11 subsets and is divided into seen and unseen sets to comprehensively evaluate the model's segmentation and generalization ability in VNS scenarios. The details are shown in Tab. \ref{tab:eval-set}.

For the seen set, it includes 6 subsets, \textit{i.e.}, CAMO (250 images) \cite{camo}, COD10K (2026 images) \cite{cod10k}, Kvasir (100 images) \cite{kvasir}, ClinicDB (62 images) \cite{CVC-ClinicDB}, DIS-Dark (480 images) \cite{dis}, and Thin-Dark (1000 images) \cite{thin}. Specifically, we assess the methods on the test sets of these datasets.
Note that, the DIS-Dark and Thin-Dark are also synthesized data by CycleGAN \cite{cyclegan} as stated above.
These data belong to the same datasets as those in the training set.

For the unseen set, it includes 5 subsets: NC4K \cite{nc4k}, ColonDB \cite{ColonDB}, ETIS \cite{etis}, LIS \cite{lis}, and CDS2K \cite{csu}.
Note that these subsets are all collected from real-world scenarios, which better assess the model's performance in real-world applications.
Specifically, NC4K is a large-scale testing dataset for camouflaged object detection, comprising 4121 images. 
ColonDB and ETIS are the commonly used datasets for polyp segmentation, comprising 380 and 196 images, respectively.  
LIS is a real-world instance segmentation dataset in low-light conditions. 
We use its RGB-dark test set for our evaluation, containing 2230 images.
In addition, the unseen-set includes a novel VNS scenario not covered in the training set: the industrial defect dataset CDS2K. 
The images in CDS2K are selected from real industrial defects databases, containing positive and negative splits. We only use the positive samples with 1330 images for our evaluation. The defect regions are relatively small and have similar patterns to the background, making them highly challenging.

\subsection{Evaluation Metrics}
We employ five metrics to assess our model's performance:
\begin{itemize}
    \item \textbf{Intersection over Union} (IoU) is a widely used metric to measure segmentation accuracy. It measures the overlap between predicted and ground truth segmentation masks.
    \item \textbf{Boundary Intersection over Union} (BIoU) \cite{biou} is an extension of the traditional IoU metric. It focuses on the boundaries of the segmented objects, providing a more sensitive assessment of boundary accuracy.
    \item \textbf{Enhanced-alignment measure ($E_{\phi}$)} \cite{e-measure} is a binary foreground evaluation metric. This metric is naturally suited for local and global similarities between binary maps. Note that we report mean $E_{\phi}$ in the experiments. 
    \item \textbf{Weighted F-measure} ($F^{w}_{\beta}$)\cite{f-measure} is based on F-measure. It incorporates spatial information, giving more importance to accurately segmenting regions near the object's boundaries and less importance to the background. 
    \item \textbf{Average Precision} (AP) summarizes the precision-recall curve into a single number, capturing the trade-off between precision and recall across different threshold values. It is used to evaluate the performance of instance segmentation in our experiments.
\end{itemize}

\begin{table*}[ht]
\centering
\renewcommand\arraystretch{1.25}
\setlength{\abovecaptionskip}{0pt}%
\setlength{\belowcaptionskip}{5pt}%
\caption{{Performance comparison on the \textit{\textbf{Eval-Unseen-Set}} of VNS-SEG. The data in the unseen-set all come from realistic scenarios. Our method consistently outperforms other competitors, highlighting its potential for extensive real-world application.}}
\setlength{\tabcolsep}{1.35 mm}{
\begin{tabular}{c|cccc|cccc|cccc|cccc|ccc}\toprule[1.pt]
     \multirow{2}{*}{{Method}} & \multicolumn{4}{c|}{{NC4K}} & \multicolumn{4}{c|}{{ColonDB}} &\multicolumn{4}{c|}{{ETIS}} & \multicolumn{4}{c|}{{CDS2K}} & \multicolumn{3}{c}{LIS} \\
\cline{2-20}
                               & IoU & BIoU & $E_{\phi}$ & $F_{\beta}^{w}$ & IoU & BIoU  & $E_{\phi}$ & $F_{\beta}^{w}$ & IoU & BIoU   & $E_{\phi}$ & $F_{\beta}^{w}$ & IoU & BIoU  & $E_{\phi}$ & $F_{\beta}^{w}$ & AP & AP$_{50}$ & AP$_{75}$ 
                            \\
\toprule[1.pt]
\multicolumn{20}{c}{\textit{\textbf{Point-Prompting-Based Evaluation}}} \\
\hline
SAM \cite{sam}
& .713   & .553   & .822   & .685
& .568   & .408   & .724   & .481
& .613   & .526   & .765   & .529
& .415   & .363   & .620   & .400 
& -      & -      & -\\
HQ-SAM \cite{samhq} 
& .794   & .625   & .915   & .817 
& .722   & .549   & .864   & .744 
& .731   & .633   & .876   & .751 
& .558   & .488   & .795   & .617 
& -      & -      & -\\
HQ-SAM-F \cite{samhq} & .799   & .634   & .904   & .792 
& .680   & .488   & .833   & .611 
& .713   & .601   & .862   & .654 
& .483   & .419   & .715   & .461 
& -      & -      & -\\
\rowcolor{lightgray!40} 
VNS-SAM (Ours)  
& \textbf{.834}   & \textbf{.683}   & \textbf{.946}   & \textbf{.868} 
& \textbf{.810}   & \textbf{.624}   & \textbf{.953}   & \textbf{.871} 
& \textbf{.852}   & \textbf{.745}   & \textbf{.971}   & \textbf{.881} 
& \textbf{.618}   & \textbf{.544}   & \textbf{.858}   & \textbf{.677} 
& -      & -      & -\\
\hline
\multicolumn{20}{c}{\textit{\textbf{Noise-Box-Prompting-Based Evaluation}}} \\
\hline
SAM \cite{sam}
& .760   & .592   & .882   & .784
& .822   & .616   & .954   & .874
& .851   & .732   & .961   & .886
& .593   & .517   & .815   & .644
& .298   & .570   & .281\\
HQ-SAM \cite{samhq} 
& .780   & .608   & .912   & .813 
& .826   & .639   & .955   & .875 
& .816   & .709   & .942   & .866 
& .551   & .480   & .811   & .633 
&.303    & .568   & .286\\
HQ-SAM-F \cite{samhq} & .801   & .640   & .921   & .828 
& .841   & .650   & .963   & .888 
& .866   & \textbf{.765}   & .972   & .896 
& .601   & .526   & .837   & .659
& .308   & .575   & .289 \\
\rowcolor{lightgray!40} 
VNS-SAM (Ours) & \textbf{.810}   & \textbf{.657}   & \textbf{.933}   & \textbf{.847} 
& \textbf{.842}   & \textbf{.657}   & \textbf{.964}   & \textbf{.895}
& \textbf{.873}   & .764   & \textbf{.979}   & \textbf{.907} 
& \textbf{.617}   & \textbf{.543}   & \textbf{.848}   & \textbf{.679}
& \textbf{.318}   & \textbf{.594}   & \textbf{.304}\\
\hline
\multicolumn{20}{c}{\textit{\textbf{GT-Box-Prompting-Based Evaluation}}} \\
\hline
SAM \cite{sam}
& .780   & .614   & .893   & .802
& .830   & .628   & .959   & .881
& .856   & .741   & .964   & .891
&   .601 &.527      & .818        & .648
& .439   & .784   & .433
\\
HQ-SAM \cite{samhq} & .792   & .623   & .917   & .822 
& .829   & .647   & .956   & .877 
& .828   & .724   & .946   & .874 
& .560   & .491   & .815   & .640 
& .437   & .773   & .428 \\
HQ-SAM-F \cite{samhq} & .815   & .659   & .928   & .839 
& .847   & .668   & .968   & .892 
& .876   & .778   & .976   & .903
& .606   & .536   & .839   & .664 
&.445 &.779 &.437\\ 
\rowcolor{lightgray!40} 
VNS-SAM (Ours) & \textbf{.830}    & \textbf{.685}   & \textbf{.941}   & \textbf{.860} 
& \textbf{.857}   & \textbf{.684}   & \textbf{.969}   & \textbf{.904} 
& \textbf{.888}   & \textbf{.790}   & \textbf{.982}   & \textbf{.918} 
& \textbf{.626}   & \textbf{.559}   & \textbf{.851}   & \textbf{.687} 
& \textbf{.461}            & \textbf{.787}            & \textbf{.457} \\
\toprule[1.pt]
\end{tabular}}
\label{tab:unseen-set}
\vspace{-4mm}
\end{table*}

\section{Experiments}
In this section, we comprehensively evaluate the proposed VNS-SAM on the VNS-SEG benchmark, including seen-set and unseen-set evaluations. 
We also perform zero-shot instance segmentation on the general COCO~\cite{coco} benchmark.
We first describe the implementation details in Section \ref{implementation-detail}. Then we compare VNS-SAM with the baseline and other competitors in Section \ref{main-exper}.
We conduct ablation studies in Section \ref{sec:ablation}. After that, more experiments and further analysis of the VNS-SAM and VNS-SEG are illustrated in Section \ref{sec: further-ana}. Finally, we conduct quantitative visualizations in Section \ref{sec: vis}.

\subsection{Experiment Details}
\label{implementation-detail}

\textbf{{Implementation Details.}}
During the training stage, the VNS-SAM is trained on the proposed VNS-SEG for 12 epochs on 4$\times$ 4090 GPUs, taking only 4 hours. 
The Adam optimizer is used with an initial learning rate of 0.001 (drops by 10$\times$ at 10 epochs) and a batch size of 16.
Unless otherwise stated, we default to using the ViT-L-based model in experiments.

During the inference stage, we follow the same pipeline of SAM but use the mask prediction from the VNS-mask token as the results for VNS objects. We comprehensively evaluate the performance under various prompts, including box, noise box, and random points. For box-prompting-based evaluation, we use the ground truth mask to generate the ground truth box and input it as the box prompt. For noise-box-prompting-based evaluation, the noise-box is generated by adding noise to
the GT box as the prompt input, following [26]. In our
experiments, the noise scale is set to 0.1 by default.
For point-prompting-based evaluation, we randomly sample several points from the ground truth masks and use them as the input prompt. In our experiments, the number of random points is set to 10 by default.

\begin{table}[t!] 
\centering
\renewcommand\arraystretch{1.1}
\setlength{\abovecaptionskip}{0pt} 
\setlength{\belowcaptionskip}{5pt}
\caption{{Comparison with other domain-specific SAM variants in the medical area. VNS-SAM achieves superior generalization, with further gains when equipped with an adapter for task-specific adaptation.}}
\setlength{\tabcolsep}{0.5 mm}{
\begin{tabular}{l|cc|cc|cc|cc|cc}
\toprule[0.9pt]
 \multirow{2}{*}{{Method}}  &\multicolumn{2}{c|}{{ClinicDB}}  & \multicolumn{2}{c|}{{Kvasir}}  &\multicolumn{2}{c|}{{ColonDB}}  &\multicolumn{2}{c|}{{ETIS}} &\multicolumn{2}{c}{{Avg.}}   \\
   &{{IoU}} & {{BIoU}} &{{IoU}} & {{BIoU}} &{{IoU}} & {{BIoU}} &{{IoU}} & {{BIoU}}   &{{IoU}} & {{BIoU}}\\       
\toprule[0.9pt]
SAM \cite{sam} & .788 & .574 & .725 & .476  & .795 & .567 & .838 & .703 & .787 & .580  \\
SAM-Med2D \cite{sam-med2d}  &.857	&.681  & \textbf{.866}	&.628	&.825	&.614	&.781	&.646	&.832	&.642\\
MedSAM \cite{medsam} 	&.851	&.696  &.858	& \textbf{.692}	&.830	&.638	&.840	&.738	&.845	&.691 \\
\rowcolor{lightgray!40}
VNS-SAM (Ours)  & .853	&.698  &.847	&.657	&.853	&.666	&.862	&.757	&.854	&.695 \\
\rowcolor{lightgray!40}
{+Encoder Adapter}  & {\textbf{.876}}	& {\textbf{.731}} & {.858} & {.664} & {\textbf{.874}}	& {\textbf{.708}}	& {\textbf{.890}}	&{\textbf{.799}}	& {\textbf{.875}}	& {\textbf{.726}} \\
\toprule[0.9pt]
\end{tabular}}
\label{tab:medical}
\end{table}

\textbf{Low-light datasets Synthesis. }
We first train a CycleGAN \cite{cyclegan} on paired LOL \cite{lol} datasets collected from real scenes for 100 epochs with an initial learning rate of 0.0002, which dropped at 50 epochs. After that, the pretrained CycleGAN is used to transform our selected datasets into low-light datasets.

\begin{table}[t!]
\centering
\renewcommand\arraystretch{1.1}
\setlength{\abovecaptionskip}{0pt} 
\setlength{\belowcaptionskip}{5pt}
\caption{{Ablation study of each component in VNS-SAM. The original SAM and finetuned SAM (\textit{FT}-Decoder) are used as the baseline.
VNS-T indicates the VNS-tokens, and DPE indicates the dual-level prediction enhancement. Each introduced module positively impacts the performance.}}  
\setlength{\tabcolsep}{0.9 mm}{
\begin{tabular}{c|c|c|c|cccc|c}
\toprule[0.9pt]
 \multirow{2}{*}{Module} & \multicolumn{2}{c|}{METI} & \multirow{2}{*}{NSFM} & \multirow{2}{*}{IoU} & \multirow{2}{*}{BIoU} & \multirow{2}{*}{$E_{\phi}$} & \multirow{2}{*}{$F_{\beta}^{w}$}  & Learnable \\ 
       & VNS-T  & DPE &  &  &  &   &   & Params \\
\toprule[0.9pt]
SAM \cite{sam} & -  &  - &  -&.720 &.561 &.825 &.736  & - \\
{\textit{FT}-Decoder} & -  &  - &  - & {.777}  & {.619} & {.893} & {.793} & {15.0 M} \\
\hline
\multirow{4}{*}{VNS-SAM}& \checkmark &   &   &.797  & .643  & .908  & .818 & 1.1 M \\ 
& \checkmark  &  \checkmark &   & .809  &.663  & .917  &.833  & 6.1 M \\
&\checkmark  & & \checkmark  & .814 &.667 &.917 & .830 & 4.8 M \\ 
\rowcolor{lightgray!40}
&  \checkmark  & \checkmark  & \checkmark  &\textbf{.821} &\textbf{.684} &\textbf{.929} &\textbf{.850} & 9.8 M\\
\toprule[0.9pt]
\end{tabular}}
\label{tab:ablation}
\end{table}

\begin{table}[t!]
\centering
\renewcommand\arraystretch{1.2}
\setlength{\abovecaptionskip}{0pt}%
\setlength{\belowcaptionskip}{5pt}%
\caption{Ablation study of the VNS-tokens. VNS-MT and VNS-ET indicate VNS-mask token and VNS-edge token, respectively. VNS-ET positively contributes to the performance, especially for BIoU, showing its effectiveness. }
\setlength{\tabcolsep}{2.4mm}{
{
\begin{tabular}{c|c|cccc}
\toprule[0.9pt]
 & Module &{IoU} & {BIoU} & $E_{\phi}$ & $F^{w}_{\beta}$ \\
\toprule[0.9pt]  
 \multirow{2}{*}{w/o NSFM} & w VNS-MT&.795 & .632 &.900 &.809  \\
 & w VNS-MT+VNS-ET &  \textbf{.805} & \textbf{.660} & \textbf{.911} & \textbf{.825} \\ \hline
 \multirow{2}{*}{w NSFM} &w VNS-MT & .806 &.660 &.903 &.818  \\
 & w VNS-MT+VNS-ET &\textbf{.821} & \textbf{.684} & \textbf{.929} & \textbf{.850}  \\
\toprule[0.9pt]
\end{tabular}}}
\label{tab:ablation-mtl}    
\vspace{-3mm}
\end{table}

\subsection{Performance Comparisons}
\label{main-exper}
In this experiment, we evaluate the performance of VNS-SAM on the VNS-SEG benchmark, including seen-set evaluation in Tab. \ref{tab:seen-set} and unseen-set evaluation in Tab. \ref{tab:unseen-set}. 
Three different prompts are used to comprehensively assess the model's performance for interactive segmentation.
Besides the original SAM, we also compare our method with HQ-SAM, which is an advanced variant of the original SAM. Additionally, we finetune HQ-SAM on our VNS-SEG following the same setting in \cite{samhq}, referred to as HQ-SAM-F to more comprehensively validate the effectiveness of VNS-SAM.

\textbf{{Performance on the Eval-Seen-Set of VNS-SEG.}}
In Tab. \ref{tab:seen-set}, we evaluate the performance of our VNS-SAM on the eval-seen-set of VNS-SEG, comprising six subsets: CAMO, COD10K, Kvasir, ClinicDB, DIS-Dark, and Thin-Dark.
{To assess our model’s robustness across various scenarios, we consider three types of prompts: the commonly used GT-box prompt, along with low-quality prompts such as point prompt and noisy-box prompt. As in real-world interactive segmentation applications, prompts may not always accurately enclose the object like the GT-box.
The results clearly demonstrate that VNS-SAM significantly outperforms the baseline SAM across all six subsets and all three types of prompts. Notably, VNS-SAM excels in point-prompting-based evaluations, where it surpasses the baseline by over 20 points on the ClinicDB (0.602 \textit{vs} 0.833) and DIS-Dark (0.594 \textit{vs} 0.780) subsets, highlighting its ability to handle low-quality input effectively. While HQ-SAM shows some improvements over SAM, its performance remains suboptimal in VNS scenarios. Even after fine-tuning on the VNS-SEG dataset, HQ-SAM-F shows some improvements, but its performance remains limited, indicating that it does not adequately capture subtle non-salient features. In contrast, VNS-SAM consistently outperforms both SAM and HQ-SAM-F, demonstrating its superior ability to adapt to the challenges posed by non-salient and low-quality prompts. This further validates the effectiveness of our proposed approach in real-world interactive segmentation tasks.}

\textbf{{Performance on the Eval-Unseen-Set of VNS-SEG.}}
In Tab. \ref{tab:unseen-set}, we evaluate the zero-shot performance of VNS-SAM on the eval-unseen-set of VNS-SEG. 
Notably, the data in the unseen-set all come from the real world, which better evaluates the model's performance in practical applications. 
There is also a novel VNS scenario not present in the training set, \textit{i.e.}, the industrial defect scenario. Overall, the unseen-set is particularly challenging due to its diversity of data from the real world. 
From the results, VNS-SAM consistently outperforms other methods. These results highlight the strong zero-shot generalization capabilities of our VNS-SAM and underscore its potential for extensive real-world applications.

\textbf{Comparison with Domain-Specific SAM Variants.}
In Tab. \ref{tab:medical}, we compare our proposed method with domain-specific variants of SAM, including SAM-Med2D \cite{sam-med2d} and MedSAM \cite{medsam}, both of which represent state-of-the-art approaches in the medical imaging domain.  The experiments are conducted on four polyp segmentation datasets \cite{kvasir, CVC-ClinicDB, ColonDB, etis}. Our VNS-SAM achieves a higher average IoU (0.854) than both SAM-Med2D (0.832) and MedSAM (0.845), demonstrating strong generalization ability to VNS scenarios.  Furthermore, we incorporate a learnable adapter \cite{samadapter} into the encoder of VNS-SAM (similar to SAM-Med2D) for more powerful task-specific adaptation. With this, our model achieves an average IoU of 0.875, surpassing all domain-specific baselines and further validating the flexibility and scalability of our approach.

\begin{table}[t!] 
\centering
\renewcommand\arraystretch{1.1}
\setlength{\abovecaptionskip}{0pt} 
\setlength{\belowcaptionskip}{5pt}
\caption{{Analysis of the design of mask and edge predictions layers.}}
\setlength{\tabcolsep}{0.8 mm}{
\begin{tabular}{lc|cccc|cccc}
\toprule[0.9pt]
 \multirow{2}{*}{{Method}} & Learnable & \multicolumn{4}{c|}{{Eval-Seen-set}} &\multicolumn{4}{c}{{Eval-Unseen-set}} \\
  & Params &IoU & BIoU & $E_{\phi}$ & $F^{w}_{\beta}$ & IoU & BIoU & $E_{\phi}$ & $F^{w}_{\beta}$ \\      
\toprule[0.9pt]
EPL\&EPL & 8.7 M &.820	& .670	&.928	&.844	&.790 &.663		&.932	&.831 \\
MPL\&MPL & 10.9 M &\textbf{.821} &\textbf{.684}	&.928	&.848	&\textbf{.802}	&\textbf{.681}	&\textbf{.936}	&\textbf{.844} \\
\rowcolor{lightgray!40}
MPL\&EPL &  9.8 M &\textbf{.821} &\textbf{.684} &\textbf{.929}	&\textbf{.850}	&.800 &.680 &\textbf{.936}	&.842\\
\toprule[0.9pt]
\end{tabular}}
\label{tab:ablation-EPL-MPL}
\end{table}

\begin{table}[t!] 
\centering
\renewcommand\arraystretch{1.1}
\setlength{\abovecaptionskip}{0pt} 
\setlength{\belowcaptionskip}{5pt}
\caption{{Analysis of the token fusion method. Our method outperforms other alternative approaches. ``TF'' indicates token fusion, ``CA'' indicates Cross-Attention, and ``DG'' indicates Dynamic Gating.}}
\setlength{\tabcolsep}{1 mm}{
\begin{tabular}{lc|cccc|cccc}
\toprule[0.9pt]
 \multirow{2}{*}{{Method}} & Learnable & \multicolumn{4}{c|}{{Eval-Seen-set}} &\multicolumn{4}{c}{{Eval-Unseen-set}} \\
  & Params &IoU & BIoU & $E_{\phi}$ & $F^{w}_{\beta}$ & IoU & BIoU & $E_{\phi}$ & $F^{w}_{\beta}$ \\      
\toprule[0.9pt]
w/o TF & 9.7 M & .809	&.675	&.927	&.848	&.791	&.671	&.932	&.840 \\
CA & 9.9 M &.818	&.677	&.926	&.844	&.791	&.669	&.930	&.834 \\
DG  & 10.2 M &.820	&.682	&\textbf{.929}	&\textbf{.850}	&\textbf{.802}	&\textbf{.681}	&\textbf{.936}	&.841 \\
\rowcolor{lightgray!40}
Ours &  9.8 M &\textbf{.821}	&\textbf{.684}	&\textbf{.929}	&\textbf{.850}	&.800	&.680	&\textbf{.936}	&\textbf{.842}\\
\toprule[0.9pt]
\end{tabular}}
\label{tab:ablation-tf}
\end{table}

\begin{table}[t!] 
\centering
\renewcommand\arraystretch{1.1}
\setlength{\abovecaptionskip}{0pt} 
\setlength{\belowcaptionskip}{5pt}
\caption{{Detailed ablation of the NSFM module. ``WD'' indicates Haar wavelet decomposition, ``HF'' and ``LF'' indicate high-frequency and low-frequency components, respectively.}}
\setlength{\tabcolsep}{1.3 mm}{
\begin{tabular}{l|cccc|cccc}
\toprule[0.9pt]
 \multirow{2}{*}{{Method}} & \multicolumn{4}{c|}{{Eval-Seen-set}} &\multicolumn{4}{c}{{Eval-Unseen-set}} \\
   &IoU & BIoU & $E_{\phi}$ & $F^{w}_{\beta}$ & IoU & BIoU & $E_{\phi}$ & $F^{w}_{\beta}$ \\       
\toprule[0.9pt]
w/o WD &.816	&.679	&.924	&.850	&.786	& .663	&.932	&.840 \\
w only HF &.814	&.674	&.920	&.841	&.790	&.670	&.933	& \textbf{.842} \\
w only LF & .814	&.675	&.922	&.843	&.795	&.674	&.933	&.841\\
\rowcolor{lightgray!40}
Ours NSFM &\textbf{.821} &\textbf{.684} &\textbf{.929} & \textbf{.859} &  \textbf{.800}	&\textbf{.680}	&\textbf{.936} &\textbf{.842} \\
\toprule[0.9pt]
\end{tabular}}
\label{tab:ablation-nsfm}
\end{table}

\begin{table}[t!]
\centering
\renewcommand\arraystretch{1.1}
\setlength{\abovecaptionskip}{0pt} 
\setlength{\belowcaptionskip}{5pt}
\caption{{Comparison of different finetuning (\textit{FT}) strategies. $FT$-Decoder and $FT$-Token indicate finetuning the entire decoder and finetuning the output token, respectively.}}
\setlength{\tabcolsep}{1. mm}{
\begin{tabular}{l|cc|cc|cc|cc}
\toprule[0.9pt]
 \multirow{2}{*}{{\textit{FT}-Strategy}} & \multicolumn{2}{c|}{{Seen-set}} &\multicolumn{2}{c|}{{Unseen-set}}  &\multicolumn{2}{c|}{{COCO}} & Learnable \\
   &{{IoU}} & {{BIoU}} &{{IoU}} & {{BIoU}} &{{IoU}} & {{BIoU}} & Params\\       
\toprule[0.9pt]
SAM & .720 & .561 &.768 & .628  & .812 &.707  & -\\
{\textit{FT}-Encoder\&Decoder} & {\textbf{.827}} & {.682} & {.771} & {.650} & {.443}  & {.334} & {1191 M} \\
\textit{FT}-Decoder &.777 &.619  & .699 & .559 & .626  & .507 & 15.0 M\\
\textit{FT}-Token   & .787  & .625 & .786  & .651  & .811  & .710  & 2.1 M \\
\rowcolor{lightgray!40}
VNS-SAM (Ours) & .821 & \textbf{.684} & \textbf{.800} &\textbf{.680} & \textbf{.816} & \textbf{.711} & 9.8 M\\
\toprule[0.9pt]
\end{tabular}}
\label{tab:ablation-tuning}
\end{table}

\subsection{Ablation Study}
\label{sec:ablation}
In this section, we conduct extensive ablation experiments and further discussions about the components of VNS-SAM to illustrate its effectiveness.

\textbf{Effect of Each Component.}
We conducted an ablation study to examine the effect of each designed component, including the VNS-tokens (VNS-T), dual-level prediction enhancement (DPE), and the Non-Salient Feature Mining module (NSFM). The results are shown in Tab. \ref{tab:ablation}. We report the average performance on the eval-seen-set and the additional parameters introduced by each technique.  
{The original SAM and the fine-tuned SAM are used as the baseline,  achieving an average IoU of 0.720 and 0.777, respectively, across the six datasets.
\textbf{i) VNS-T}: To encourage the decoder to learn VNS characters, we first add a pair of VNS-tokens, including a VNS-mask token and a VNS-edge token. After incorporating and fine-tuning VNS-tokens on the task data, a significant performance improvement is observed, outperforming the fine-tuned SAM by about 2 points with fewer learnable parameters {(1.1 M \textit{vs} 15.0 M)}.
\textbf{ii) DPE}: Building upon the previous step, we improve the SAM's decoder to a dual-level enhanced decoder, which allows both decoding layers to consistently learn subtle discriminative features, thereby improving the decoder’s ability to understand non-salient characteristics. DPE consistently improves IoU, BIoU and $F_{\beta}^{w}$ by more than 1 point.
 \textbf{iii) NSFM}: Furthermore, we introduce NSFM to mine the useful low-level features from the highly optimized image encoder to enrich the representation of the prediction layer. 
With the help of NSFM, the performance is consistently improved (the last two rows) and the final IoU achieves 0.821, with an improvement of more than 10 points compared to the baseline SAM and 5 points to fine-tuned SAM. Notably, the NSFM is lightweight, with only {3.7 M} parameters introduced. 
Overall, the results reported in Tab. \ref{tab:ablation} show that each introduced module positively impacts VNS-SAM’s performance.}

\begin{table}[t!] 
\centering
\renewcommand\arraystretch{1.1}
\setlength{\abovecaptionskip}{0pt} 
\setlength{\belowcaptionskip}{5pt}
\caption{{Performance comparison on the COCO set. The COCO dataset is partitioned into salient (COCO-S) and non-salient (COCO-NS) subsets using the VNS-score. VNS-SAM consistently outperforms SAM, particularly in non-salient scenarios, demonstrating its robustness in challenging conditions.}}
\setlength{\tabcolsep}{2 mm}{
\begin{tabular}{l|cc|cc|cc}
\toprule[0.9pt]
 \multirow{2}{*}{Method} & \multicolumn{2}{c|}{COCO-all} & \multicolumn{2}{c|}{COCO-S}  &\multicolumn{2}{c}{{COCO-NS}} \\
   &{{IoU}} & {{BIoU}} &{{IoU}} & {{BIoU}} &{{IoU}} & {{BIoU}} \\
\toprule[0.9pt]
SAM \cite{sam}&.755	&.642	&.763	&.666	&.732	&.572 \\
\rowcolor{lightgray!40}
VNS-SAM (Ours) & \textbf{.775}	&\textbf{.661}	& \textbf{.778}	& \textbf{.681}	& \textbf{.765}	& \textbf{.604}\\
\toprule[0.9pt]
\end{tabular}}
\label{tab:ablation-coco}
\end{table}

\begin{table*}[h]
\centering
\renewcommand\arraystretch{1.1}
\caption{{Performance comparison between SAM and VNS-SAM across different ViT backbones. VNS-SAM consistently outperforms the baseline across different backbones and datasets with only increasing a small number of extra parameters.}}
\setlength{\tabcolsep}{1.9 mm}{
\begin{tabular}{cccccccccccccccc}\toprule[1.3pt]
     \multirow{2}{*}{{Backbone}} & \multirow{2}{*}{{Method}}  &  \multicolumn{4}{c}{{\textit{\textbf{Eval-Seen-Set}}} } & \multicolumn{4}{c}{{\textit{\textbf{Eval-Unseen-Set}}}} &\multicolumn{4}{c}{{\textit{\textbf{COCO-all}}}} & \multicolumn{2}{c}{Model Params (MB)} \\
\cmidrule(r){3-6}  \cmidrule(r){7-10} \cmidrule(r){11-14} \cmidrule(r){15-16}
                           &     & IoU & BIoU & $E_{\phi}$ & $F_{\beta}^{w}$ & IoU & BIoU & $E_{\phi}$ & $F_{\beta}^{w}$ & IoU & BIoU & $E_{\phi}$ & $F_{\beta}^{w}$  & Total  & Learnable\\
\toprule[1.3pt]
\multirow{2}{*}{ViT-B}  
& SAM \cite{sam} & .652 & .486 & .757 & .641  &.735 &.584 &.868 &.742 & .784 & .670 & .927 & .850 & 358  & 358 \\
& \cellcolor{lightgray!40} VNS-SAM   & \cellcolor{lightgray!40} \textbf{.794} &\cellcolor{lightgray!40} \textbf{.643} &\cellcolor{lightgray!40} \textbf{.912} &\cellcolor{lightgray!40} \textbf{.817} & \cellcolor{lightgray!40} \textbf{.782} & \cellcolor{lightgray!40} \textbf{.652} &\cellcolor{lightgray!40} \textbf{.927} &\cellcolor{lightgray!40} \textbf{.822} &\cellcolor{lightgray!40} \textbf{.798} & \cellcolor{lightgray!40} \textbf{.689} &\cellcolor{lightgray!40} \textbf{.951} & \cellcolor{lightgray!40} \textbf{.866}  & \cellcolor{lightgray!40} 367.4 & \cellcolor{lightgray!40} 9.4 \\
\hline
\multirow{2}{*}{ViT-L}
& SAM \cite{sam} &.720 &.561 &.829 & .737  & .768 & .628 & .909 & .805  & .812 & .707 & .955 & \textbf{.881}  & 1191  & 1191 \\
& VNS-SAM   \cellcolor{lightgray!40} & \textbf{.821} \cellcolor{lightgray!40} &\textbf{.684} \cellcolor{lightgray!40}& \textbf{.929} \cellcolor{lightgray!40}& \textbf{.850} \cellcolor{lightgray!40}&\textbf{ .800} \cellcolor{lightgray!40}& \textbf{.680} \cellcolor{lightgray!40}& \textbf{.944} \cellcolor{lightgray!40}& \textbf{.853} \cellcolor{lightgray!40}& \textbf{.816} \cellcolor{lightgray!40}& \textbf{.711} \cellcolor{lightgray!40}& \textbf{.956} \cellcolor{lightgray!40}& \cellcolor{lightgray!40} \textbf{.881}  & \cellcolor{lightgray!40} 1200.8  &\cellcolor{lightgray!40} 9.8\\
\hline
\multirow{2}{*}{ViT-H} 
& SAM \cite{sam} &.716 &.566 &.830 &.741  &.767 &.633 &.910 &.808 & .812 & .710 & \textbf{.956} & .878  & 2446  & 2446\\
& \cellcolor{lightgray!40} VNS-SAM  & \cellcolor{lightgray!40} \textbf{.833} & \cellcolor{lightgray!40} \textbf{.697} & \cellcolor{lightgray!40} \textbf{.936} & \cellcolor{lightgray!40} \textbf{.858}  & \cellcolor{lightgray!40} \textbf{.800} & \cellcolor{lightgray!40} \textbf{.677} & \cellcolor{lightgray!40} \textbf{.934} & \cellcolor{lightgray!40} \textbf{.838}  & \cellcolor{lightgray!40} \textbf{.814} & \cellcolor{lightgray!40} \textbf{.714} & \cellcolor{lightgray!40} \textbf{.956} & \cellcolor{lightgray!40} \textbf{.879}  & \cellcolor{lightgray!40} 2456.2 &\cellcolor{lightgray!40} 10.2 \\
\toprule[1.3pt]
\end{tabular}}
\vspace{-5pt}
\label{tab:various-backbone}
\end{table*}

\begin{table*}[h]
\centering
\renewcommand\arraystretch{1.1}
\setlength{\abovecaptionskip}{0pt}%
\setlength{\belowcaptionskip}{5pt}%
\caption{{The effect of our unified VNS-SEG dataset. NC4K, ETIS, and LIS are unseen datasets for camouflaged, polyp, and low-light object segmentation, respectively. Using VNS-SEG for training consistently achieves excellent results across multiple tasks and outperforms the results trained on specialized datasets in each task.}}
\setlength{\tabcolsep}{2.1 mm}{
\begin{tabular}{ccccccccccccc}\toprule[1.pt]
    \multirow{2}{*}{{Train Set}} &\multirow{2}{*}{{VNS Character}} & \multicolumn{4}{c}{{NC4K}} & \multicolumn{4}{c}{{ETIS}} & \multicolumn{3}{c}{LIS}\\
                &     & IoU & BIoU & $E_{\phi}$ & $F_{\beta}^{w}$  & IoU & BIoU & $E_{\phi}$ & $F_{\beta}^{w}$ & AP & AP$_{50}$ & AP$_{75}$ \\
\toprule[1.pt]
Baseline      &    -       & .780  &.614  &  .893  & .802 & .856 & .741 & .964 &.891 & .439 & .784 & .433 \\ \hline
COD10K+CAMO  & Camouflage  & {.825}   & {.679}   &{.937}   & {.854}  & .857   & .756   & .965   & .889  &.445 & {.785} & .437 \\
Kvasir+ClinicDB  & Polyp  & .730   & .530   & .890   & .768  & {.850}   & {.748}   & {.958}   & {.885} & .399 & .708 & .392 \\ 
DIS-Dark+Thin-Dark+FSS-Dark & Low-light & .787  & .634   & .920   & .828  & .867   & .769 & .972   & .906 &{.459} & .786 & {.455}\\
\rowcolor{lightgray!40}
VNS-SEG & Unified VNS characters  & \textbf{.830} &\textbf{.685} &\textbf{.941} & \textbf{.860} & \textbf{.888} & \textbf{.790} & \textbf{.982} & \textbf{.918} & \textbf{.461} & \textbf{.787} & \textbf{.457} \\
\toprule[1.pt]
\end{tabular}}
\vspace{-10pt}
\label{tab:effect-vns-seg}
\end{table*}

\textbf{Ablation of META.}
Tab.~\ref{tab:ablation-mtl} presents the dissected ablation study on the VNS-tokens, illustrating the respective impacts of the  VNS-mask token (VNS-MT) and VNS-edge token (VNS-ET) on the model’s performance. 
It is evident that the VNS-edge token plays a crucial role in enhancing model performance, particularly for BIoU, with improvements of 2.8 points (without NSFM) and 2.4 points (with NSFM).

\textbf{Analysis of the design of MPL and EPL.}
To verify the rationality of the structural design of MPL and EPL, we conduct detailed ablation experiments that are shown in Tab. \ref{tab:ablation-EPL-MPL}. Specifically, (a) ``EPL\&EPL'': we remove the SAM pre-trained features from MPL, making it structurally consistent with EPL. (b) ``MPL\&MPL'': we incorporate SAM pre-trained features into EPL, aligning its structure with MPL. 
The overall performance of ``EPL\&EPL'' is inferior to the original design while the performance of ``MPL\&MPL'' is comparable to the original design, but introduces more parameters. Our approach offers a simpler structure while achieving better performance.

{\textbf{Analysis of Token Fusion Strategies.} 
After each decoder layer, the VNS-edge token is integrated with the VNS-mask token to explicitly aggregate the edge representation. 
We explore various integrating strategies, including (a) cross-attention, (b) dynamic gating, and (c) our element-wise addition with linear fusion. The results are shown in Tab. \ref{tab:ablation-tf}. It shows that the token integration operation effectively improves IoU and BIoU by approximately 1 point. These three strategies achieve comparable performance, but our method requires fewer parameters and adopts a more straightforward form.}

{\textbf{Ablation of NSFM Module.}  
In NSFM, we employ Haar wavelet decomposition \cite{haar} to separate features into four frequency bands and select low-frequency and high-frequency components for further processing to enhance segmentation robustness. To verify the rationality of our approach, we conducted a detailed ablation study, as shown in Tab. \ref{tab:ablation-nsfm}. The results indicate that removing the wavelet decomposition (WD) leads to a performance drop, while utilizing only high-frequency (HF) or only low-frequency (LF) components results in suboptimal performance. 
Our NSFM module achieves the best results across all metrics, demonstrating its effectiveness in leveraging both high- and low-frequency features for improved segmentation performance.
}

\textbf{Comparison with Other Finetuning Strategies. }
{In Tab.~\ref{tab:ablation-tuning}, we compare our method with other finetuning strategies, including finetuning SAM's encoder \& decoder (\textit{FT}-Encoder \& Decoder), decoder (\textit{FT}-Decoder), and finetuning its output mask token (\textit{FT}-Token). 
The performance is evaluated on the VNS-SEG and COCO datasets.
It can be observed that:  
\textbf{i)} Finetuning the encoder \& decoder, or only the decoder, enhances performance on the VNS-SEG seen-set by better capturing non-salient characteristics. However, this comes at the cost of catastrophic forgetting, leading to a dramatic performance drop on COCO.}
\textbf{ii)} Compared to finetuning the entire decoder, finetuning the output mask token effectively improves the performance. However, it is still limited in visually non-salient scenarios. 
\textbf{iii)} Our approach effectively exploits SAM’s low-level features to boost the learning of VNS characters, bringing a large performance improvement and remaining powerful zero-shot segmentation ability.

\begin{figure*}[ht]
\centering
\includegraphics[width=0.85 \textwidth]{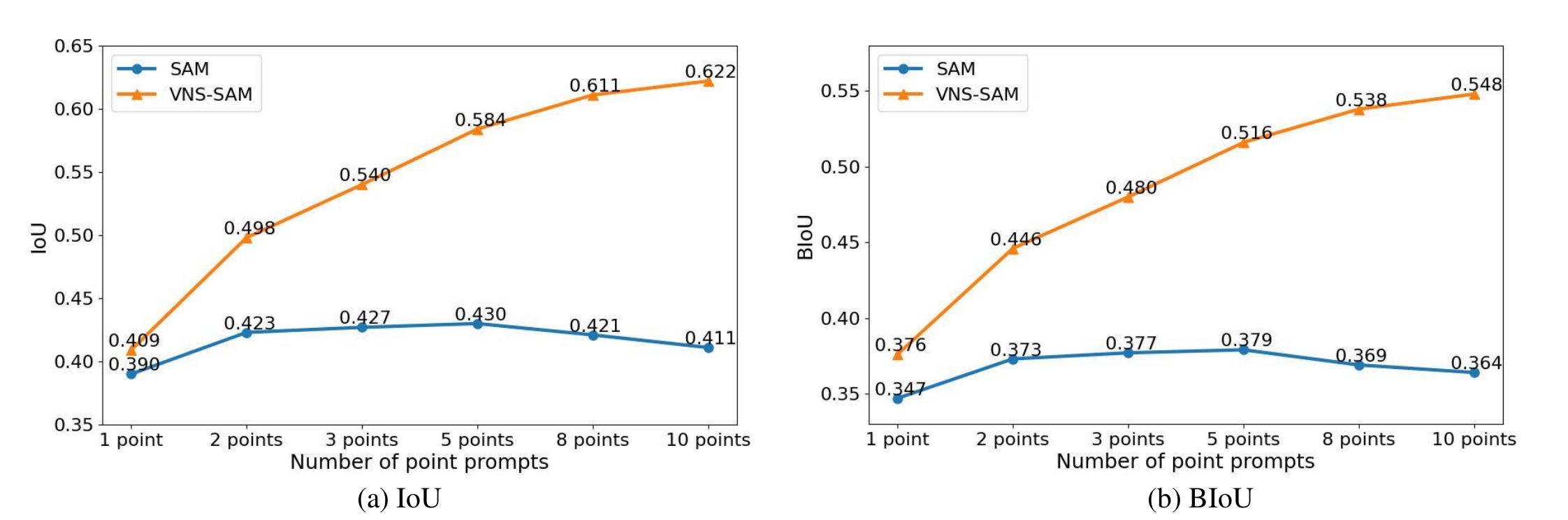}
\caption{
Performance comparison of interactive segmentation with varying quantities of input points on the unseen subset CDS2K. VNS-SAM consistently outperforms SAM across a range of point counts, demonstrating a more significant improvement.
}
\label{fig:point_com}
\end{figure*}

\begin{figure*}[h]
\centering
\includegraphics[width=0.95 \textwidth]{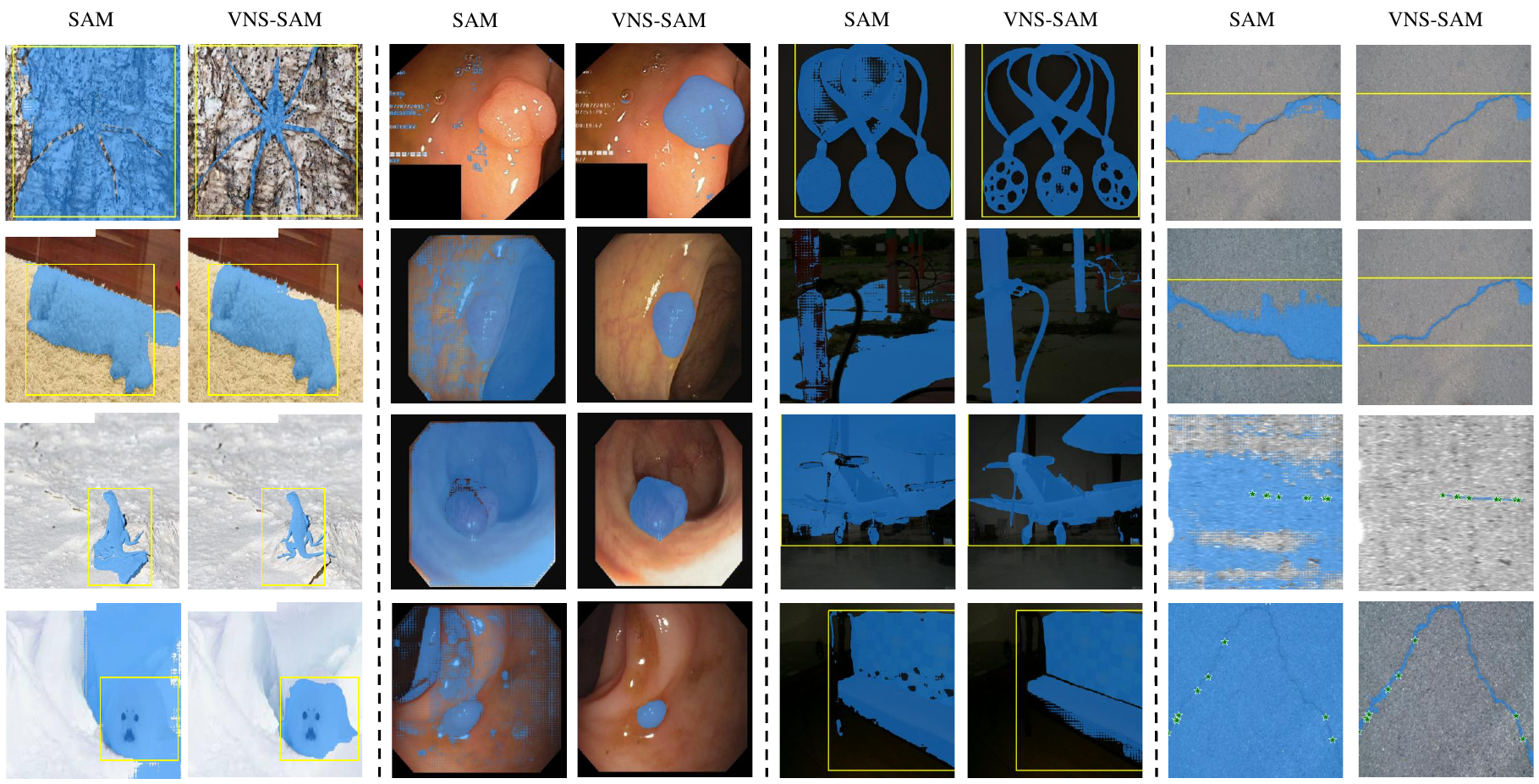}
\caption{
{Visual comparisons of segmentation results between SAM and VNS-SAM. When facing challenging VNS scenarios, SAM fails to accurately distinguish between the foreground and background, resulting in incorrect segmentation. In contrast, our VNS-SAM is more robust towards these scenarios.
}}
\label{fig:result-vis}
\end{figure*}

\subsection{Further Analysis and Discussion}
\label{sec: further-ana}

{\textbf{Performance on General COCO dataset.}
To further validate our method in more general non-salient scenes, we design a Visually Non-Saliency Score (VNS-score) that quantifies the image’s non-saliency (more details are in the Appendix \ref{vns-score}) and extract a non-salient sub-dataset using the VNS-score within the COCO dataset. 
We calculate the score for each object-image pair and extract a non-salient subset (COCO-NS) with a threshold of 0.7, while the remaining are categorized as the salient subset (COCO-S).
Benchmarking results on these curated subsets are shown in Tab. \ref{tab:ablation-coco} that underscore the efficacy of our approach. On the challenging COCO-NS subset, VNS-SAM achieves an IoU of 76.5\% and a BIoU of 60.4\%, surpassing SAM by 3.3\% and 3.2\%, respectively. Importantly, VNS-SAM maintains competitive performance on the COCO-S subset (77.8\% IoU \textit{vs} SAM's 76.3\%), confirming that its improvements in non-salient scenes do not come at the expense of general segmentation performance.}

\textbf{{Comparison across Different Backbones.}}
In Tab. \ref{tab:various-backbone}, we conduct a thorough comparison between SAM and VNS-SAM across various ViT \cite{vit} backbones, including ViT-Base (ViT-B), ViT-Large (ViT-L), and ViT-Huge (ViT-H). 
We comprehensively assess the models on the seen and unseen sets of the VNS-SEG and COCO datasets. 
The performance of the seen, unseen, and COCO datasets are reported. In addition, the total and learnable parameters of models are also included.
These results demonstrate that VNS-SAM consistently outperforms SAM with significant margins on various sizes of backbones and different datasets.  In terms of model size, ViT-B, ViT-L, and ViT-H-based VNS-SAM only increase 2.5\%, 0.8\%, and 0.4\% parameters, respectively.

\textbf{Effect of VNS-SEG.}
In Tab. \ref{tab:effect-vns-seg}, we compared the results of single-task data training with unified data training, clearly demonstrating the advantages of the VNS-SEG dataset. 
For camouflaged, polyp, and low-light object segmentation tasks, we use the commonly used training sets of COD10K+CAMO, Kvasir+ClinicDB, and DIS-Dark+Thin-Dark+FSS-Dark for training the model respectively.
We use three unseen datasets for zero-shot evaluation, \textit{i.e.}, NC4K, ETIS, and LIS.
Notably, using VNS-SEG for training consistently achieves excellent results across multiple tasks and outperforms the results trained on specialized datasets in each task. 
This indicates that the unified VNS-SEG dataset enables the model to learn more robust non-salient characters, which is superior to the previous single-task dataset for training.

\begin{figure}[t]
\centering
\includegraphics[width=0.5 \textwidth]{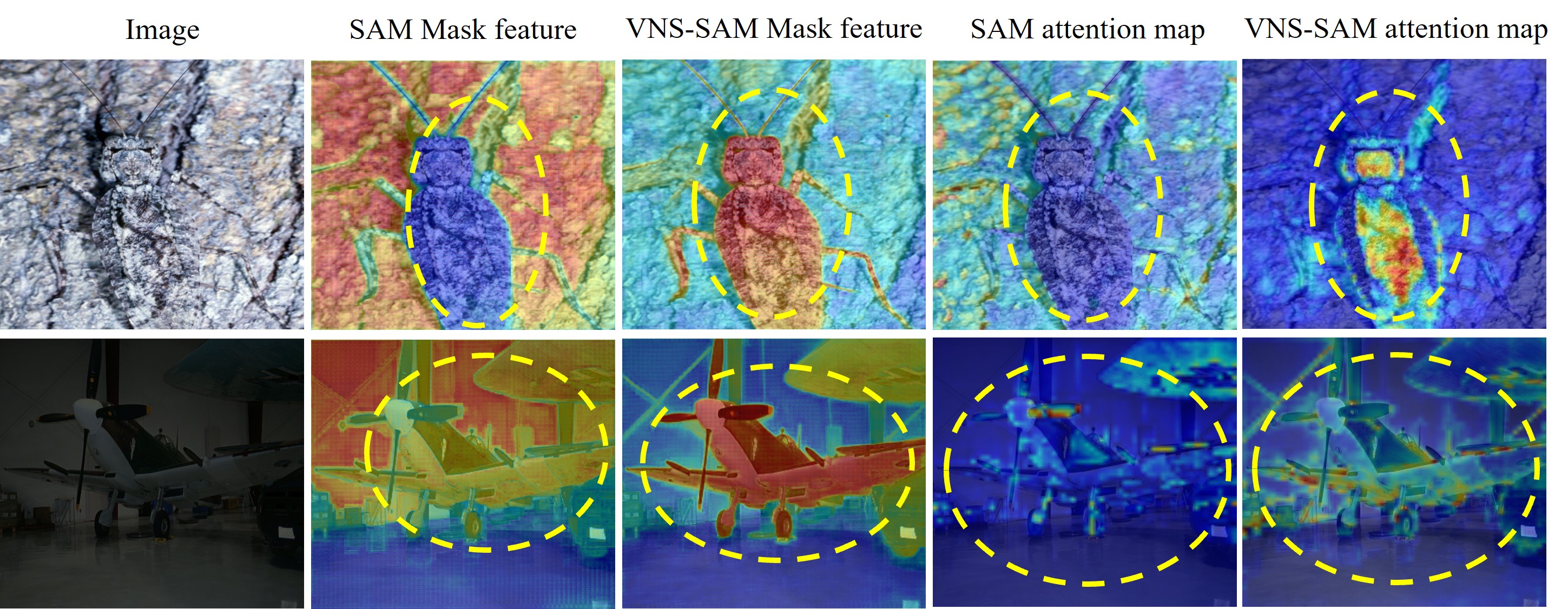}
\caption{
Visual comparisons of mask features (the feature maps multiplied with the output tokens in the prediction layer) and attention maps (cross-attention map in the final decoder layer of the output token) between SAM and VNS-SAM. 
VNS-SAM showcases accurate activation of the target areas and boundaries, while SAM, due to its lack of ability to recognize non-salient characteristics, exhibits confusion between the foreground and background.
}
\label{fig:feature-map}
\end{figure}

\begin{table}[t] 
\centering
\renewcommand\arraystretch{1.}
\setlength{\abovecaptionskip}{0pt}%
\setlength{\belowcaptionskip}{5pt}%
\caption{{Computational requirements of SAM, HQ-SAM, and VNS-SAM.}}
\setlength{\tabcolsep}{3 mm}{
\begin{tabular}{ccccc}\toprule[1.pt]
     \multirow{2}{*}{{Method}}  & \multicolumn{2}{c}{Params} & \multirow{2}{*}{FLOPs} & Inference\\
    &  Learnable  & Total   &  &  FPS \\
\toprule[1.pt]
SAM & 1191 M & 1191 M & $\approx$1550 G  & 7.3 \\
HQ-SAM & 5.1 M & 1196.1 M & $\approx$1553 G & 7.1 \\
\rowcolor{lightgray!40}
VNS-SAM & 9.8 M & 1200.8 M & $\approx$1559 G & 7.0 \\
\toprule[1.pt]
\end{tabular}}
\label{tab:speed}
\end{table}

\textbf{Point-based Interactive Segmentation Comparison.}
Fig. \ref{fig:point_com} presents the interactive segmentation performance of VNS-SAM and SAM using point prompts. This comparison assesses VNS-SAM and SAM with a range of input point numbers on the unseen subset CDS2K.
VNS-SAM consistently outperforms SAM across different numbers of point prompts (from 1 point to 10 points).
Note that as the prompt contains less ambiguity (with more input points), the relative performance improvement becomes more significant. 
This indicates the powerful segmentation capability of VNS-SAM.

{\textbf{Comparison of computational requirements.}
As shown in Tab. \ref{tab:speed}, VNS-SAM introduces only a marginal increase in total parameters compared to HQ-SAM (1200.8M vs 1196.1M) and FLOPs (1559G vs 1553G). Despite this slight overhead, VNS-SAM achieves substantially higher segmentation performance, demonstrating a superior performance–efficiency trade-off. The inference speed remains comparable (7.0 FPS vs. 7.1 FPS), confirming that our approach is both effective and practical for real-world deployment.}

\subsection{Visualization}
\label{sec: vis}
In this part, we present some visualization results and qualitatively compare our method with SAM.

\textbf{Segmentation Results Visualization.}
In Fig. \ref{fig:result-vis}, we present the visualized segmentation results of SAM and our VNS-SAM on the evaluation set of VNS-SEG. 
We can observe that, due to the challenging VNS characters, SAM struggles to segment these objects accurately, resulting in serious detail missing and erroneous background prediction, showing its limitations. 
In contrast, our VNS-SAM can precisely segment the inconspicuous objects in VNS scenarios, demonstrating its robust perception ability towards various VNS characters.

\textbf{Feature Visualization.}
In Fig.~\ref{fig:feature-map}, we provide an illustrative comparison of the mask feature maps (the second and third columns) and cross-attention maps (the fourth and fifth columns) of the last decoder layer between SAM and VNS-SAM.
The mask features come from the final mask prediction layer of the decoder, and the cross-attention maps come from the last token-to-image layer corresponding to the SAM's output mask token and our VNS mask token. 
It can be observed that VNS-SAM showcases accurate activation of the target areas and boundaries, while SAM, due to its lack of ability to recognize non-salient characteristics, exhibits confusion between the foreground and background. 
This demonstrates VNS-SAM's enhanced ability to distinguish subtle discriminative regions and details, which is crucial for effective segmentation under non-salient conditions.

\section{Conclusion}
In this paper, we investigate the issue of SAM's performance degradation when facing scenarios with low contrast between foreground and background, which we refer to as visually non-salient scenarios. 
To address this issue, we propose VNS-SAM to enhance SAM's perception of VNS scenarios while preserving its original zero-shot generalizability. 
We achieve this by effectively exploiting SAM’s low-level features through two effective and efficient designs: the Mask-Edge Token Interactive decoder and the Non-Salient Feature Mining module. 
From the data perspective, we establish the unified VNS-SEG that includes various VNS scenarios, in contrast to the previous single-scenario dataset. 
VNS-SEG is used to enable the model to learn robust non-salient features and comprehensively assess the model's performance in VNS scenarios. 
Extensive experiments are conducted to demonstrate the superior performance of VNS-SAM, highlighting its potential for broad real-world applications.
Additionally, the performance on the seen and unseen sets of VNS-SEG establishes a new standard for VNS segmentation. 
In terms of future research, we hope the constructed VNS-SEG dataset will inspire more powerful segmentation models suitable for VNS scenarios.

\appendix
\subsection{{Visually Non-Saliency Score}}
\label{vns-score}
{To further analysis, we design a \textbf{Visually Non-Saliency Score} (VNS-score) that quantifies the image's non-saliency. It is calculated from two aspects: the contrast between foreground and background, and the clarity of object boundaries. Specifically, the calculation of the foreground-background contrast $C_{fb}$ comprehensively takes into account two key factors: color contrast and texture contrast. Color contrast reflects the difference in color between the foreground and background, while texture contrast reflects the difference in their texture features. The color contrast is measured by calculating the difference between the color mean vectors $\mu_{fg}^{LAB}$ and $\mu_{bg}^{LAB}$ of the foreground and background regions in the LAB color space \cite{lab-color}. Texture contrast is calculated based on the Gray-Level Co-Occurrence Matrix (GLCM) \cite{glcm}. 
The contrast of the foreground region $\text{C}_{fg}^{GLCM}$and the contrast of the background region $\text{C}_{bg}^{GLCM}$ are obtained, respectively.
\begin{equation}
    C_{fb}=\frac{1}{2}(\|\mu_{fg}^{LAB}-\mu_{bg}^{LAB}\|+\|\text{C}_{fg}^{GLCM}-\text{C}_{bg}^{GLCM}\|).
\end{equation}
The boundary clarity $B$ is used to measure the clarity of the object boundaries in an image. It is defined as:
\begin{equation}
    B = \frac{\text{Mean}(\|\nabla I_{\text{edge}}\|)}{255},
\end{equation}
where $\|\nabla I_{\text{edge}}\|$ represents the gradient magnitude calculated using the Sobel operator, specifically within the object boundary regions. A value of $B$ close to 0 suggests significant blurriness of the object boundaries.}

{Finally, the VNS-score is obtained by a weighted sum of the $C_{fb}$ and the $B$, as:
\begin{equation}
    \text{VNS-score}=1-\frac{1}{2}(C_{fb}+B).1
\end{equation}
}

\begin{table}[h] 
\centering
\renewcommand\arraystretch{1.1}
\setlength{\abovecaptionskip}{0pt} 
\setlength{\belowcaptionskip}{5pt}
\caption{{The mean and standard deviation of the original datasets, synthetic low-light datasets, and real low-light datasets.}}
\setlength{\tabcolsep}{1.5 mm}{
\begin{tabular}{l|cc|cc|cc}
\toprule[0.9pt]
 \multirow{2}{*}{{Dataset}} & \multicolumn{2}{c|}{{Original}} & \multicolumn{2}{c|}{{Synthetic low-light}}  &\multicolumn{2}{c}{{{Real low-light}}} \\
   & {Mean} & {SD} & {Mean} & {SD} &{{Mean}} & {SD} \\
\toprule[0.9pt]
 {Mean/SD}  & {131.82}	& {58.72}	& {11.13}	& {7.81} & {8.31} & {10.3} \\
\toprule[0.9pt]
\end{tabular}}
\label{tab:mean-sd}
\end{table}

\subsection{{Realism of the Synthetic Datasets}}
{We computed the mean and variance of the images for original datasets (DIS, Thin, and FSS), the corresponding synthetic datasets (DIS-Dark, Thin-Dark, and FSS-Dark) generated by CycleGAN, and real low-light datasets LIS \cite{lis}. The results are shown in Tab. \ref{tab:mean-sd}. 
Compared to the original datasets, the synthetic images achieved the mean and standard deviation (SD) much closer to those of the real low-light LIS dataset (11.13 vs 8.31 and 7.81 vs 10.3). This demonstrates that CycleGAN-generated data effectively captures the statistical properties of non-salient scenarios, even if absolute photorealism is not achieved.}

\bibliographystyle{IEEEtran}
\bibliography{main}

\vfill

\end{document}